\documentclass{ieeeaccess}
\usepackage{cite}
\usepackage{amsmath,amssymb,amsfonts}
\usepackage{algorithm}
\usepackage{algorithmic}
\usepackage{graphicx}
\usepackage{subfigure}
\usepackage{textcomp}
\usepackage{adjustbox}
\usepackage{amsthm}

\usepackage{url}

\newtheorem{lemma}{Lemma}

\newtheorem{prop}{Proposition}

\DeclareMathOperator*{\argmax}{arg\,max}
\DeclareMathOperator*{\argmin}{arg\,min}
\DeclareMathOperator*{\st}{\text{subject to}}

\def\BibTeX{{\rm B\kern-.05em{\sc i\kern-.025em b}\kern-.08em
    T\kern-.1667em\lower.7ex\hbox{E}\kern-.125emX}}
\begin{document}
\history{This paper has been accepted by IEEE Access.}
\doi{10.1109/ACCESS.2022.3209243}

\title{ASK: Adversarial Soft k-Nearest Neighbor Attack and Defense}
\author{\uppercase{Ren Wang}\authorrefmark{1,5}\IEEEmembership{Member, IEEE}, 
\uppercase{Tianqi Chen}\authorrefmark{2}, \uppercase{Philip Yao}\authorrefmark{1}, \uppercase{Sijia Liu}\authorrefmark{3}, \uppercase{Indika Rajapakse}\authorrefmark{4} and \uppercase{Alfred Hero}\authorrefmark{1} \IEEEmembership{Life Fellow, IEEE}}
\address[1]{Department of Electrical Engineering and Computer Science, University of Michigan, Ann Arbor, MI 48109 USA}
\address[2]{Department of Statistics, University of Michigan, Ann Arbor, MI 48109 USA}
\address[3]{Department of Computer Science and Engineering, Michigan State University, East Lansing, MI 48824 USA}
\address[4]{Department of Computational Medicine and Bioinformatics, University of Michigan, Ann Arbor, MI 48109 USA}
\address[5]{Department of Electrical and Computer Engineering, Illinois Institute of Technology, Chicago, IL 60616 USA}
\tfootnote{This work was supported in part by the United States Department of Defense, Defense Advanced Research Projects Agency (DARPA) under Grant HR00112020011, by the Army Research Office (ARO) under grant W911NF-15-1-0479, and by the Department of Energy National Nuclear Security Administration under grant DE-NA0003921.} 

\markboth
{Author \headeretal: Preparation of Papers for IEEE TRANSACTIONS and JOURNALS}
{Author \headeretal: Preparation of Papers for IEEE TRANSACTIONS and JOURNALS}

\corresp{Corresponding authors: Ren Wang (e-mail: rwang74@iit.edu); Indika Rajapakse (e-mail: indikar@umich.edu); Alfred Hero (e-mail: hero@eecs.umich.edu).}

\begin{abstract}
K-Nearest Neighbor (kNN)-based deep learning methods have been applied to many applications due to their simplicity and geometric interpretability. However, the robustness of kNN-based deep classification models has not been thoroughly explored and kNN attack strategies are underdeveloped. In this paper, we first propose an \underline{A}dversarial \underline{S}oft \underline{k}NN (ASK) loss for developing more effective kNN-based deep neural network attack strategies and designing better defense methods against them. Our ASK loss provides a differentiable surrogate of the expected kNN classification error. It is also interpretable as it preserves the mutual information between the perturbed input and the in-class-reference data. 
We use the ASK loss to design a novel attack method called the ASK-\underline{At}tac\underline{k} (ASK-Atk), which shows superior attack efficiency and accuracy degradation relative to previous kNN attacks on hidden layers. We then derive an ASK-\underline{Def}ense (ASK-Def) method that optimizes the worst-case ASK training loss. Experiments on CIFAR-10 (ImageNet) show that (i) ASK-Atk achieves $\geq 13\%$ ($\geq 13\%$) improvement in attack success rate over previous kNN attacks, and (ii) ASK-Def outperforms the conventional adversarial training method by $\geq 6.9\%$ ($\geq 3.5\%$) in terms of robustness improvement. Relevant codes are available at \url{https://github.com/wangren09/ASK}.
\end{abstract}


\titlepgskip=-15pt

\maketitle

\section{Introduction}
\label{sec: intro}
\PARstart{T}{he} K-Nearest Neighbor (kNN) classifier is a simple prototype classification algorithm that has been widely used in classification tasks \cite{cover1967nearest,tang2020predicting,balsubramani2019adaptive}. Combined with deep neural networks (DNNs), kNN has also been successfully applied to a broad range of important learning tasks, e.g., Remote Sensing Image Retrieval \cite{ye2019new}, machine translation \cite{khandelwal2021nearest}, speaker verification \cite{khan2020vector}, and face recognition \cite{nakada2017acfr}. Nonetheless, DNN classifiers have been found to be vulnerable against adversarial attacks \cite{kurakin2018adversarial,SZS14}, which generate small perturbations on benign examples to flip class predictions. Although researchers have recently demonstrated that kNN-based deep learning methods are effective in defending against adversarial attacks on DNNs \cite{papernot2018deep,wang2022rails}, and can be used in more general data quality testing applications \cite{bahri2020deep}, as compared to DNN's, the intrinsic adversarial robustness of kNN-based deep classifiers has been less well explored.

\begin{figure*}[h]
  \centering
  \includegraphics[trim=60 0 60 0,clip,width=.99\textwidth]{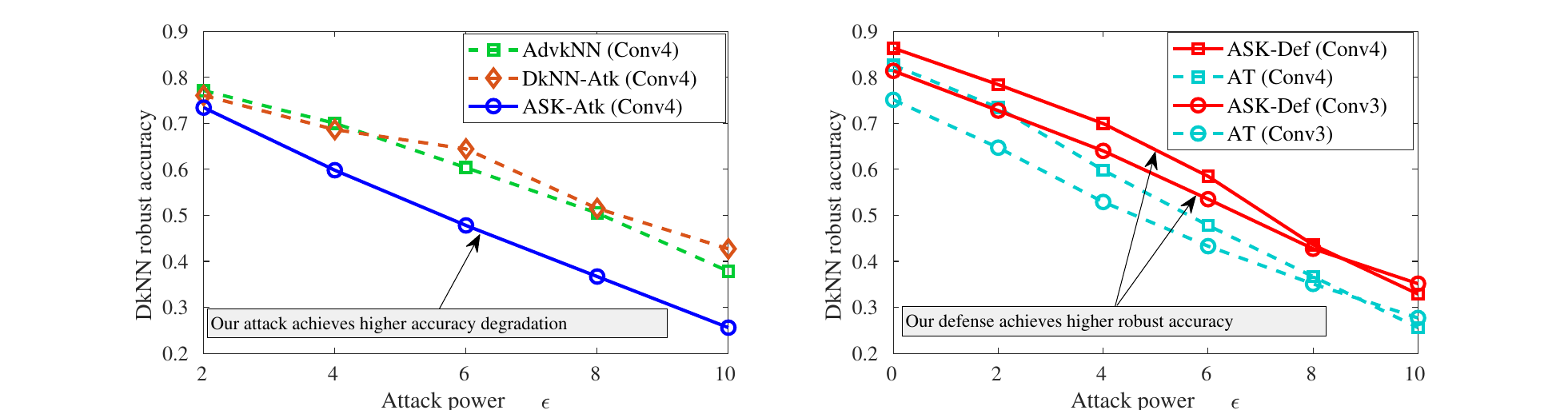}
  \caption{{\bf Proposed attack (ASK-Atk on left panel) and proposed defense (ASK-Def on right panel) outperform previous methods in terms of impact on accuracy. } All accuracy curves correspond to attacks or defenses on the base classifier, a Deep k-Nearest Neighbor (DkNN) \cite{papernot2018deep}, with VGG16 CNN architecture, trained on CIFAR-10. Left: ASK-Atk achieves higher accuracy degradation (higher attack success rate) than the existing DkNN-Atk \cite{sitawarin2019robustness} and AdvKnn \cite{li2019advknn}
  attacks on the adversarially trained  \cite{madry2017towards} model. Right: Model trained using the proposed defense (ASK-Def) is more resilient to the ASK-Atk than the adversarially trained baseline model for defending convolutional layer blocks 3 (Conv 3) and 4 (Conv 4) of VGG16. We refer readers to Section~\ref{sec: experiments} and Appendix~\ref{sec: app5}, \ref{sec: app6} to see more results of ASK-Atk and ASK-Def.}
  \label{fig: att_defense}
\end{figure*}

Previously proposed strategies for evaluating kNN-based deep learning methods on image classification tasks include: the Deep kNN attack (DkNN-Atk), which is based on a heuristic process to approximate the kNN \cite{sitawarin2019robustness}; and the adversarial KNN attack (AdvKnn) \cite{li2019advknn}, which uses a neural network to approximate the kNN distribution. As shown in  Figure~\ref{fig: att_defense}, neither work particularly well on the CIFAR-10 DNN model trained by the adversarial training (AT) \cite{madry2017towards}. 

In this paper, we develop a more effective optimization-based attack strategy on kNN-type deep classifiers using a novel loss function, called the Adversarial Soft kNN (ASK) loss. ASK loss better approximates the kNN's probability of classification error as compared to the previously introduced loss functions. We show that the ASK loss is bounded by an information-theoretic measure: a mutual information measure between the perturbed input and benign examples, providing intuition for attack power transfer and the ability to defend against hidden-layer attacks. As our major contribution, we develop a more effective attack on kNN-based deep classifiers, called the ASK-Attack (ASK-Atk), that maximizes the ASK loss. When applied to the kNN on DNN structures, our proposed ASK-attack targets its hidden layers. Therefore, ASK-attack can also be leveraged to evaluate the robustness of features extracted in a hidden layer.

We then derive a more robust defense against kNN-based attacks on hidden layers, called the ASK-Defense (ASK-Def), that minimizes the maximum ASK in the robust training process. Previous robust training methods have largely focused on robustifying DNN prediction \cite{madry2017towards,zhang2019theoretically}. To the best of our knowledge, the proposed ASK-defense is the first method to defend against kNN attacks on hidden layers of DNNs via robust training. The left and right panels of Figure~\ref{fig: att_defense} demonstrates the effectiveness of both the proposed ASK-Atk and ASK-Def methods, respectively. Importantly, ASK-Def achieves an average $7\%$ robustness with training complexity comparable to conventional adversarial training.

\paragraph{Contributions.} Our principal contributions are the following:

\begin{itemize}
    \item A novel Adversarial Soft kNN (ASK) loss is introduced that approximates the kNN's probability of classification error in the adversarial setting, and establish that it guides the design of effective attack and defense strategies through the perspective of preserving mutual information between the perturbed input and the in-class-reference data.
    \item A new attack strategy, the ASK-Attack (ASK-Atk), is presented that maximizes the ASK loss. ASK-Attack provides a principled and effective way to evaluate the robustness of kNN-based deep learning methods in both white-box and gray-box settings, and can be leveraged to evaluate the robustness of features extracted in a hidden layer. On CIFAR-10 and Imagenette we show that the proposed ASK-Atk improves on the kNN attacks of \cite{sitawarin2019robustness} and \cite{li2019advknn}: ASK-Atk achieves $\geq 13\%$ additional degradation in accuracy rates.
    \item A new defense strategy, the ASK-Defense (ASK-Def), is presented to robustify model weights under various kNN-based attacks targetting hidden layers through minimizing the maximum ASK loss. On CIFAR-10 (Imagenette) the proposed defense improves the robustness of the DkNN over conventional adversarial training \cite{madry2017towards} by $\geq 6.9\%$ ($\geq 3.5\%$) on the single selected layers and $4.8\%$ ($1.3\%$) on the combined layers. 
\end{itemize}

\paragraph{Background and comparisons.}
Non-parametric methods like kNN have been widely applied to different tasks, e.g., neural language models \cite{khandelwal2021nearest,khan2020vector} and the reasoning task of correspondence classification \cite{plotz2018neural}. Recently, kNN has received renewed attention for its potential to improve robustness. In \cite{wang2018analyzing} the authors proposed a robust $1$-NN classifier that removes a subset corresponding to the oppositely labeled nearby points in the training data. Multi-layer kNN's \cite{papernot2018deep,wang2022rails} provide robust classification against adversarial attacks on DNNs. The baseline method we use to evaluate the ASK-Atk and ASK-Def is the Deep k-Nearest Neighbor (DkNN) \cite{papernot2018deep}. In a DkNN the kNN classifier is embedded into selected layers of a DNN, producing an output class decision by majority vote among the kNN classifiers: 
\begin{align}\label{eq: dknn}
    \begin{array}{ll}
y_{\tiny \rm{DkNN}} =   \argmax_c \sum_{l \in \mathcal{L}} p_l^c(\mathbf x),  ~~~ c \in [C],
    \end{array}
\end{align}
where $l$ is the $l$-th layer of a DNN and $\mathcal{L}$ is the set of the selected layers. Here $p_l^c(\mathbf x)$ is the confidence score assigned to class $c$ predicted by the kNN in layer $l$ for input $\mathbf x$. $[C]$ denotes the set $\{1,2,\cdots,C\}$, where $C$ is the number of classes. Recent work shows that the DkNN can also be used to remove the mislabeled training data \cite{bahri2020deep}.

While robustness of kNN-based deep classifiers against a wide range of adversarial attacks has been established, their robustness against customized attacks is less well studied. Originally designed with cosine similarity, the DkNN attack (DkNN-Atk) attempts to reduce robustness by moving examples towards incorrect classes through some heuristic steps \cite{sitawarin2019robustness}. Another attack, AdvKnn, trains a small neural network, called the deep kNN block, to approximate the kNN classifier and uses it to construct the attack \cite{li2019advknn}. Our ASK loss framework for designing adversarial attacks establishes a much stronger attack, the ASK-Atk, than these previous attacks. This is illustrated for the DkNN/VGG16 model trained on the CIFAR-10 training set in the left panel of Figure~\ref{fig: att_defense}. The results also demonstrate that the ASK loss can well approximate the kNN's probability of classification error. One recent work \cite{frosst2019analyzing} has studied a similar form of loss for representation learning with random minibatch selection. However, the loss is based on pairwise similarity measurement on a single layer and does not consider the adversarial setting. In contrast, the ASK loss is a class-wise loss under the adversarial setting and has a more general extension with a weighted summation of different layers. Moreover, the ASK loss provides interpretability of effective attack in white-box and gray-box settings. The form of ASK loss can be compared with contrastive loss functions, which has often been used in self-supervised representation learning, e.g., for data augmentation by applying operations like cropping, resizing, or rotation 
\cite{chen2020simple,sohn2016improved}. 
This paper is the first to propose a loss of such form (see \eqref{eq: knn_ask}) in the supervised and adversarial DNN setting.

\begin{figure*}[h]
  \centering
  \includegraphics[trim=0 0 0 0,clip,width=.78\textwidth]{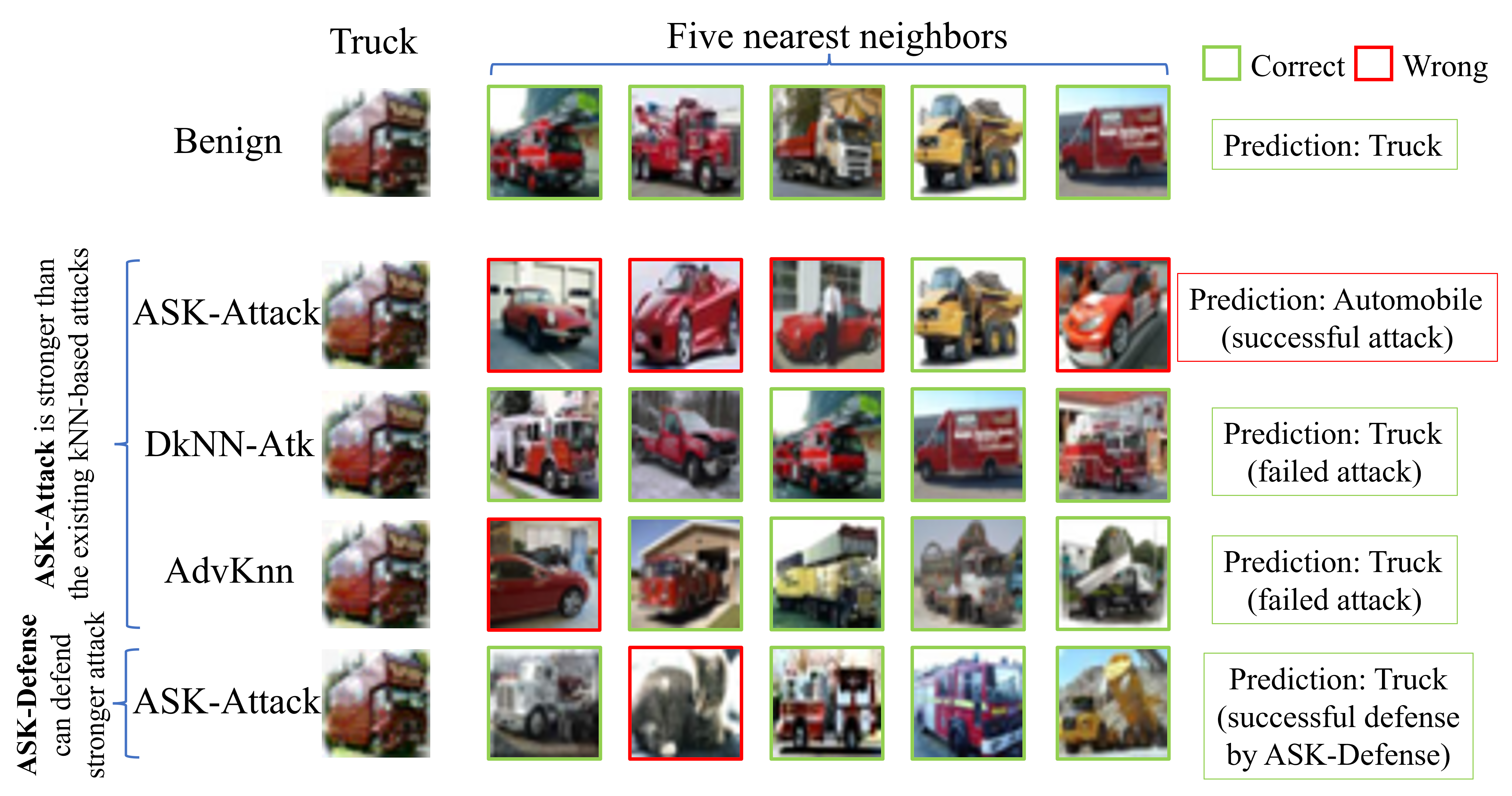}
  \caption{{\bf Representative example showing superiority of proposed ASK attack (ASK-Atk) and ASK defense (ASK-Def) as compared to previous attack strategies (DkNN-Atk and Adv-Knn)}.  When the DkNN/VGG16 predictor is trained with standard adversarial training \cite{madry2017towards} on CIFAR-10, for equal perturbation $\epsilon$, ASK-Atk (2nd row) is more successful then other attacks (3rd and 4th row). Ask-Atk maximizes the proposed ASK loss, and its perturbation of the truck image (top left) flips the classes of 4 out of 5 of the image's nearest neighbors. When DkNN/VGG16 is trained with Ask-Def, which minimizes the maximum ASK loss, it successfully defends against all 3 attacks. In particular, the ASK-Atk can only flip the class of a single nearest neighbor (5th row). }   
  \label{fig: vis_diff_methods}
\end{figure*}

Furthermore, as ASK-Atk is based on maximizing the ASK loss function, a stronger defense can be mounted by minimizing the maximum ASK loss during the training process, leading to the proposed ASK-Def. Unlike other robust training methods that defend the output layer of the DNN \cite{madry2017towards,zhang2019theoretically}, ASK-Def also defends kNN classifiers on hidden layers leading to improved robustness (right panel of Figure~\ref{fig: att_defense}). In Figure~\ref{fig: vis_diff_methods} we illustrate the advantage of ASK-Atk and ASK-Def for one of the CIFAR-10 images contributing to the curves shown in Figure~\ref{fig: att_defense}.  For a clean example of class ``truck,'' the figure shows that \textbf{(i)} ASK-Atk successfully attacks the example while the AdvKnn and DkNN-Atk fail, and \textbf{(ii)} the model using standard adversarial training  fails to defend against ASK-Atk while the model trained by ASK-Def successfully defends against the attack.

\section{Preliminary}

\paragraph{Notation.} We consider a DNN with weights $\theta$, and denote $\theta_{l}$ as weights from layer $1$ to layer $l$. $f_{\theta_{l}}(\cdot): \mathbb{R}^{d} \rightarrow \mathbb{R}^{d_{l}}$ represents the mapping from the input to the $l$-th layer feature representation. $(\mathbf x, y)$ represents a benign input-label pair. $\mathbf x^\prime = \mathbf x + \mathbf \delta$ denotes the perturbed version of $\mathbf x$. In the adversarial context, $\mathbf x^\prime$ is called an adversarial example that follows a target adversarial distribution. $(X^{l+}, Y^{l+})$ denotes a subset containing $K$ data points selected from $\mathcal{D}_{y}$ (the training data of class $y$). Similarly, $(X_c^{l-}, Y_c^{l-})$ is the subset of the training data of class $c$, where $c \in [C]$, and we denote the $\{X_c^{l-}\}_{\forall c \in [C], c\not= y}$ by $X^{l-}$. We will call $X^{l} = \{X^{l+}, X^{l-}\}$ the {\em reference data} for $\mathbf x$ at the $l$-th layer, $X^{l+}$ the {\em in-class reference data}, and $X^{l-}$ the {\em out-of-class reference data}.  
$A(\cdot,\cdot)$ is a similarity measurement between a data pair, and can be chosen from different similarity functions, e.g., cosine, inner-product, and (negative) $\ell_2$ distance. Specifically, the cosine similarity is $A(f_{\theta_{l}}; \mathbf x_1, \mathbf x_2) = -\frac{1}{\tau_l} \cos{\big(f_{\theta_{l}}(\mathbf x_1), f_{\theta_{l}}(\mathbf x_2)\big)}$, where $\tau_l$ denotes a positive (temperature) constant that rescales the similarity of the data pair in the $l$-th layer. The negative $\ell_2$ distance similarity is also used in our experiments and is defined as: $A(f_{\theta_{l}}; \mathbf x_1, \mathbf x_2) = -\frac{1}{\tau_l} \|f_{\theta_{l}}(\mathbf x_1)- f_{\theta_{l}}(\mathbf x_2)\|$.

\paragraph{Attacker's knowledge.} We consider both a white-box and a gray-box setting for the adversarial attack. In both settings, the adversary has knowledge of the model parameters, similarity metric, and selected subset of the training set used to perform the kNN search. In the white-box setting, the adversary also knows the layers used for kNN searching and the number of nearest neighbors $K$. In the gray-box setting, either the layer information or the number $K$ is not available to the attacker. We will focus on the $\ell_\infty$ attack.

\paragraph{Defender's capabilities.} In our setting, the model trainer is the defender who only has access to the training process while having no control of the model during the inference phase.

\section{Adversarial soft kNN (ASK) loss}\label{sec: ask_loss}

In this section, we will propose a loss that will be used to design effective attacks and defenses on kNN-based deep classifiers. The weights of DNNs are most commonly trained using gradient descent methods. Such methods require a smooth (differentiable or sub-differentiable) network architecture, as a function of the DNN weights, in addition to a smooth loss function. However, neural nets incorporating kNN's are not differentiable as the kNN mapping is not a smooth function of its inputs.
Besides overcoming the non-differentiability challenge, we also hope the attack can work well when the kNN implementation layers and $K$ of the kNN-DNN hybrid system are unknown to attackers, i.e., under our gray-box setting. As for the defense, it needs to be effective in facing kNN-based attacks on deep classifiers.  
Spurred by these challenges, our design of the loss function is guided by three principles: \textbf{(1)} The loss should provide a differentiable surrogate of the expected kNN classification error, and thus makes it easy to evaluate its robustness. \textbf{(2)} The loss should enable attack power transfer across layers and various choices of $K$ in our gray-box setting (without knowing the exact layers of implementing kNN and exact $K$). \textbf{(3)}  The loss needs to improve model robustness against different attacks on hidden layers. 



\paragraph{ASK loss.} We introduce a simple and differentiable Adversarial Soft kNN (ASK) loss, which is a function of the kNN classifier only through the kNN distances. For perturbed input $\mathbf x^\prime$, the \textit{ASK loss} on layer $l$ is defined as
\begin{align}\label{eq: knn_ask}
    & L_{\mathrm{ask}}^l(\theta_{l}; \mathbf x^\prime, X^{l}) = - \mathbb E_{ (X^{l}, \mathbf x^\prime) } \big[\\
    &
    \log \frac{  S(f_{\theta_{l}}; \mathbf x^\prime, X^{l+}) }{   S(f_{\theta_{l}}; \mathbf x^\prime, X^{l+})
     + \sum_{c=1,c\neq y}^C  S(f_{\theta_{l}}; \mathbf x^\prime, X_c^{l-}) }\big]\nonumber , 
\end{align}
where $S(f_{\theta_{l}}; \mathbf x^\prime, X)$ is a similarity measure between $\mathbf x^\prime$ and the data samples $X$ on layer $l$. $f_{\theta_{l}}$ acts on both $\mathbf x^\prime$ and $X$. We consider two specifications of $S(\cdot,\cdot)$: \textbf{(i)} $S(\cdot,\cdot) = \exp{\big(\frac{1}{K}\sum_{k=1}^K A(\cdot,\cdot)\big)}$ in our attack (the ASK-Attack Section); and \textbf{(ii)} $S(\cdot, \cdot) = \sum_{k=1}^K \exp{\big( A(\cdot,\cdot)\big)}$ in our defense (the ASK-Defense Section). Note that \eqref{eq: knn_ask} is still well defined when there is no DNN, i.e, $f_{\theta_{l}}$ is an identity map. 

\paragraph{Information-theoretic interpretation of ASK loss.}  The ASK loss \eqref{eq: knn_ask} can be defined at any layer $l$ and is closely related to the mutual information between $X^{l+}$ and the perturbed input $\mathbf x^\prime$. Specifically, the following proposition shows that the negative ASK loss $-L_{\rm ask}^l $ lower bounds the mutual information $\mathrm{MI} (X^{l+},  \mathbf x^\prime)$ up to an additive constant $\log (C-1)$.
\begin{prop}\label{eq: thm}
For perturbed input $\mathbf x^\prime$ that follows a target distribution,
\begin{align}\label{eq: ineq}
\mathrm{MI} (X^{l+},  \mathbf x^\prime) \geq  -L_{\rm ask}^l  + \log (C-1)
\end{align}
\end{prop}

The proof is given in Appendix~\ref{sec: app1}, which follows a similar line of argument to that used in infoNCE \cite{oord2018representation}. One can also check that the ASK losses under two forms of $S(\cdot, \cdot)$ (used for design of the attack and the defense) lower bound the mutual information. In practice, we would hope the encoding network $\theta_{l}$ is deep enough in order to provide good feature extraction. 

Using kNN search, $X^{l+}$ and $X^{l-}$ are the $K$ closest data points to the example $\mathbf x$ on layer $l$ from each respective class. The kNN classifier works well when the data points within the same class are consistent and close to each other, while data points from different classes are far apart. Proposition~\ref{eq: thm} makes it clear that the attacker and defender should take different strategies. From the defender's point of view, minimizing $L_{\rm ask}^l$ can enhance the model robustness by increasing the mutual information between the perturbed input and the in-class reference data, which helps defend against different attacks on hidden layers without knowing the exact attack forms (principle \textbf{(3)}). In contrast, an attacker would want to minimize the mutual information to create maximum ambiguity between the perturbed and unperturbed inputs which tends to make ASK loss large when there are a large number $C$ of classes. In particular, if the attacker can force the mutual information to zero, then $L_{\rm ask}^l$ can't be smaller than $\log(C-1)$. Proposition~\ref{eq: thm} 
also indicates that in the sense of minimizing the mutual information, the attack can be effective without knowing the exact layers of implementing kNN and exact $K$ (principle \textbf{(2)}). Intuitively, minimizing \eqref{eq: knn_ask} forces $(\mathbf x^\prime, X^{l+})$ to be mapped close together, while forcing pairs $(\mathbf x^\prime, X^{l-})$ further away in the embedding space induced by the $l$-th layer. The ASK loss has the interpretation as a temperature-scaled cross-entropy \cite{bengio2017deep}, which is a differentiable surrogate of the expected kNN classification error (principle \textbf{(1)}). We will consider two types of ASK loss in the following Sections: ASK loss obtained through locally searched reference data in the attack and ASK loss obtained through non-locally selected reference data with class constraints in our defense.

\section{ASK-Attack}\label{sec: attack}

The attacker's goal is to find a small perturbation $\mathbf \delta$ such that $\mathbf x^\prime = \mathbf x + \mathbf \delta$ flips the kNN classifier prediction (at the input layer or at a deep layer). The strength of the attack $\mathbf \delta$ is constrained to the $\ell_p$ ball: $\mathcal{C} = \{\mathbf \delta\ |\ \|\mathbf \delta\|_p \le \epsilon \}$. 

\paragraph{Attack strategy.}
The attacker's goal is to push the perturbed input $\mathbf x^\prime$ away from the $y$ classification region, i.e., the region in which the kNN classifier would correctly predict $y$, and move $\mathbf x^\prime$ towards some data points from other classes. Motivated by the properties of the ASK loss discussed in the previous Section, we adopt a attack strategy that maximizes the weighted summation of ASK losses, i.e., it solves the optimization problem $\arg\max_{\mathbf \delta \in \mathcal{C}} \sum_{l \in \mathcal{L}} \omega_l L_{\mathrm{ask}}^l$, or more explicitly
\begin{align}\label{knn_att_nont}
     & \arg\max_{\mathbf \delta  \in \mathcal{C}} \sum_{l \in \mathcal{L}}
    - \omega_l \cdot \\&  \log  \frac{ S(f_{\theta_{l}}; \mathbf x + \mathbf \delta, X^{l+}) 
    }{
    S(f_{\theta_{l}}; \mathbf x + \mathbf \delta, X^{l+}) + 
   \sum_{c=1,c\neq y}^C S(f_{\theta_{l}}; \mathbf x + \mathbf \delta, X_c^{l-})  
    }, \nonumber
\end{align}
where $S(f_{\theta_{l}}; \mathbf x + \mathbf \delta, X) = \exp{ \frac{1}{K}\sum_{k=1}^K A(f_{\theta_{l}}; \mathbf x + \mathbf \delta, X) }$. $\omega_l$ is a weighting coefficient applied to layer $l$, and $\sum_{l \in \mathcal{L}}\omega_l = 1$. Note that \eqref{knn_att_nont} corresponds to a non-targeted attack. For a targeted attack the strategy is to move an example $\mathbf x + \mathbf \delta$ to a target class $c_t$ that is different from class $y$. For the targeted attack we replace $\sum_{c=1,c\neq y}^C S(f_{\theta_{l}}; \mathbf x + \mathbf \delta, X_c^{l-})$ with $S(f_{\theta_{l}}; \mathbf x + \mathbf \delta, X_{c_t}^{l-})$. The target class can be randomly selected or pre-selected by calculating the average distance to $\mathbf x$. In the sequel ASK-Atk will always denote the non-targeted attack, unless otherwise specified. Results for the targeted attack are presented in Appendix~\ref{sec: app5}. \eqref{knn_att_nont} can be efficiently solved using gradient ascent with projection, with iterations defined as 
\begin{align}\label{knn_att_grad}
    \mathbf x^{\prime} \leftarrow \prod_{\mathcal{B}(\mathbf x^{\prime},\epsilon)} \big[ \mathbf x^{\prime} + \kappa {\rm{sign}}\big(\nabla_{\mathbf x^{\prime}} \sum_{l \in \mathcal{L}} \omega_l L_{\mathrm{ask}}^l(\theta_{l}, \mathbf x^{\prime}, X^{l})\big)\big] 
\end{align}
where $\prod$ is a projection operator, and $\mathcal{B}(\mathbf x^{\prime},\epsilon)$ represents the projection set that satisfies the prior constraints on the data domain (e.g., $\mathbf x^{\prime} \in [0,255]^d$ when $\mathbf x^{\prime}$ is an image vector) and the $\ell_p$ ball constraint. $\kappa$ is the step size. Note that implementing the ASK-Atk on any layer and any $K$ can affect the mutual information between the perturbed input and the in-class reference data. This enables the attack to work in the gray-box setting. More details are included in Section~\ref{sec: experiments} and Appendix~\ref{sec: app2}. Next we introduce important parameter selection strategies and computational cost.

\paragraph{Selection of $X^{l+}$ and $X^{l-}$.} The in-class reference points $X^{l+}$ for $\mathbf x$ are selected by kNN search for the $K$ data points closest to $\mathbf x$ in class $y$. The out-of-class reference points $X^{l-}$ for $\mathbf x$ are also selected using kNN search, but possibly with a different value of $K$. We consider $K$ to be the same by default. The exact kNN selection improves upon an alternative random selection method to constitute $X^{l-}$ presented in Appendix~\ref{sec: app5}.

\paragraph{Selection of $\tau_l$ and $\omega_l$.} Attacker selection of the hyperparameter $\tau_l$ that scales the similarity between the data pair is selected via cross-validation,
which only requires a one-time kNN search on a small batch. We also use cross-validation to select  $\omega_l$, which controls the power of attacking layer $l$ under the multi-layer attack. This is discussed in more detail in the Section~\ref{sec: experiments}.

\paragraph{Computational cost.} During training the bottleneck for ASK-Atk is the kNN search, which must be performed over all target samples selected from the training set and requires on the order of $O(kn \log n)$ computations.  This is a fixed setup cost that can be handled off-line by parallel or distributed computing, fast approximate kNN approximation  \cite{bewley2013advantages,rajaraman2011mining}. After training, the cost for on-line prediction is cheap, e.g., using a ball-tree to perform kNN search is only of order $O(k\log n)$. $\tau_l$ and $\omega_l$ only requires a one-time kNN search on a small batch, therefore has limited impact on the computational time.

\section{ASK-Defense}\label{sec: RT}
The proposed defense ASK-Def solves a minimax optimization problem to robustify the DkNN against attacks and, in particular, kNN-based attacks. ASK-Def is constructed in two stages corresponding to adversarial examples generation and model update, 
\begin{align}\label{eq: optimization}
    \begin{array}{ll}
 &\arg\min_\theta \sum_{(\mathbf x,y) \in \mathcal D} L_{\rm ce}(\theta, \mathbf x^{\rm{adv1}}, y)+ \\& ~~~~~~~~~~~~\lambda \sum_{l \in \mathcal{L}}L_{\rm ask}^l\big(\theta_{l}, \mathbf x^{\rm{adv2}}, X^{l}\big) \\& \st ~~ \mathbf x^{\rm{adv1}}, \mathbf x^{\rm{adv2}} := {\rm{G1}}(\mathbf x), {\rm{G2}}(\mathbf x),
    \end{array}
\end{align}
where $L_{\rm ce}$ denotes the DNN cross-entropy loss. G1 and G2 denote adversarial attack generators discussed below. The $\argmin_\theta$ operation in (\ref{eq: optimization}) performs the update that integrates the DNN cross-entropy with the ASK loss, using a penalty parameter $\lambda$ to control the balance between the two loss terms. In ASK-Def, we use $S(f_{\theta_{l}}; \mathbf x^\prime, X_c^{l}) = \sum_{k=1}^K e^{ A(f_{\theta_{l}}; \mathbf x^\prime, X_c^{l}(k)) }$ as the class similarity measure. Following the same practical way of dealing with the computational burden as in the nearest neighbor approximation work \cite{frosst2019analyzing}, we find $\{X^{l+}, X^{l-}\}$ from a reference batch, which is obtained by non-local sampling with input label-aware class constraints. 
\begin{figure}[h]
  \centering
  \includegraphics[trim=0 0 0 0,clip,width=.49\textwidth]{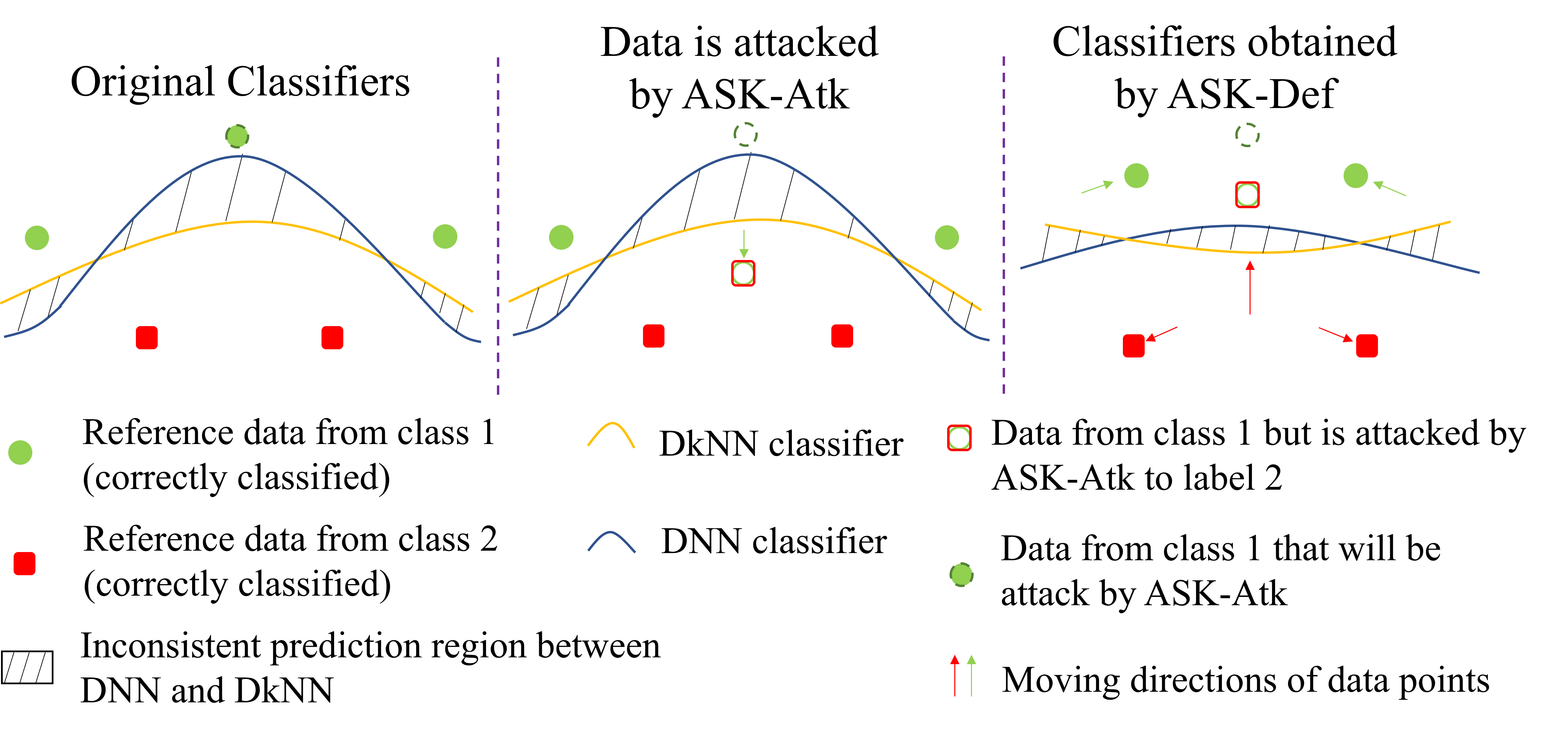}
  \caption{{\bf Conceptual diagram illustrating how proposed ASK-Def defense strategy emulates ASK-Atk and improves robustness of both the DNN and the DkNN}. Left: A DNN binary classifier has equivalent classification performance as a deeper kNN classifier but is locally more complex (higher curvature decision boundary). Benign data points (red class and green class) are well separated in the feature space. Middle: The adversarial generator ${\rm{G2}}$ generates an adversarial example that crosses both decision boundaries leading to a classification error. Right: Minimizing the ASK loss forces the adversarial example and the reference data from class 1 to be closer to each other, resulting in locally smoothed decision boundaries of the DkNN and DNN that increase attack resiliency. }
  \label{fig: mov_gen}
\end{figure}

\paragraph{Adversarial generators ${\rm{G1}}, {\rm{G2}}$.} ASK-Def uses the adversarial generators ${\rm{G1}}, {\rm{G2}}$ that generate attacks that seek to maximize two different losses 
\begin{align}\label{eq: generator}
    \begin{array}{ll}
 \noindent({\rm{G1}})~\mathbf x^{\rm{adv1}}=\mathbf x + \mathbf \delta^*, \\~~~~ \mathbf \delta^*= \arg\max_{\mathbf \delta \in \mathcal{C}} L_{\rm ce}(\theta, \mathbf x + \mathbf \delta, y), \\ ({\rm{G2}})~\mathbf x^{\rm{adv2}}=\mathbf x + \mathbf \delta_{(l)}^*, \\~~~~ \mathbf \delta_{(l)}^*= \arg\max_{\mathbf \delta \in \mathbb{\hat{
C}}} L_{\rm ask}^l(\theta_{l}, \mathbf x + \mathbf \delta, X^{l}\bigcup X^{\rm{adv2}}),
    \end{array}
\end{align}
where $X^{\rm{adv2}}$ is the adversarial counterpart of $X^{l}$. ${\rm{G1}}$ generates attacks that maximize the DNN cross-entropy loss, while ${\rm{G2}}$ generates attacks that maximize ASK loss. Thus ASK-Def reduces to conventional adversarial training (AT) \cite{madry2017towards} when $\lambda=0$ in (\ref{eq: optimization}), while it reduces to hardening against the customized ASK-Atk attack when $\lambda \gg 0$. We could also set ${\rm{G1}}$ to generate attacks by a variant of AT, e.g., TRADES \cite{zhang2019theoretically}. The pseudocode for the full ASK-Def procedure is given in Appendix~\ref{sec: app3}. 

\paragraph{Universal robustness enhancement.}
Based on Proposition~\ref{eq: thm}, minimizing the ASK loss improves the model robustness by increasing the mutual information between the perturbed input and the in-class reference data. This indicates that ASK-Def is able to increase the model resilience against different kNN-based attacks without knowing the exact attack forms. ASK-Def is also conceptually illustrated in Figure~\ref{fig: mov_gen}, in which we show how using ${\rm{G2}}$ followed by minimizing the ASK loss term help smooth local decision boundaries of DNN and (deep) kNN, increasing resiliency to attack.

\begin{table*}[ht]
\caption{{\bf ASK-Attack (ASK-Atk) outperforms AdvKnn \cite{li2019advknn} and DkNN-Atk \cite{sitawarin2019robustness} on CIFAR-10.} All the methods are applied on the same DNN model. kNN attacks and DkNN are implemented on the same layers and use either the cosine similarity or $\ell_2$ similarity in the ASK loss \eqref{eq: knn_ask}.}
\begin{center}
\label{tab: attack_comp_cifar}
\resizebox{1\textwidth}{!}{
\begin{tabular}{l||c|c|c|c|c|c}
\hline
\hline
& \multicolumn{3}{c|}{cosine similarity}  & \multicolumn{3}{c}{$\ell_2$ similarity} \\ [0.5ex] 
\hline
& \multicolumn{1}{c|}{\textbf{ASK-Atk}} & \multicolumn{1}{c|}{AdvKnn}  & \multicolumn{1}{c|}{DkNN-Atk} & \multicolumn{1}{c|}{\textbf{ASK-Atk}} & \multicolumn{1}{c|}{AdvKnn}  & \multicolumn{1}{c}{DkNN-Atk}\\ [0.5ex] 
\hline
Conv 3 & \bf{35.1\%$\pm$0.2\%} & 58.9\%$\pm$0.3\% & 60.7\%$\pm$0.44\% & \bf{32.21\%$\pm$0.16\%} & 56.55\%$\pm$0.24\% & 54.45\%$\pm$0.52\%\\
\hline
Conv 4  & \bf{36.7\%$\pm$0.18\%}  & 50.5\%$\pm$0.34\% &  50.8\%$\pm$0.39\%  & \bf{34.46\%$\pm$0.16\%}    & 53.9\%$\pm$0.26\% &  49.15\%$\pm$0.49\%\\
\hline
Conv 3,4 &   \bf{41\%$\pm$0.19\%}  & 54.1\%$\pm$0.25\% &  54.85\%$\pm$0.41\% &  \bf{39.26\%$\pm$0.17\%}   & 57.15\%$\pm$0.28\% &  54.65\%$\pm$0.5\%\\
\hline
\hline
\end{tabular}}
\end{center}
\end{table*}

\section{Experimental Results}\label{sec: experiments}
We focus on image classification tasks using CIFAR-10 and Imagenette (10 sub-classes of ImageNet) databases. VGG16 and ResNet18 are trained on CIFAR-10 and Imagenette, respectively. More details on the use of these datasets and models are given in Appendix~\ref{sec: app4}. The DkNN \cite{papernot2018deep} is used for illustration of the proposed ASK-Atk and ASK-Def  methods. We name the $i$-th convolutional layer block as Conv $i$. For VGG16 (ResNet18), Conv 4 denotes the 10-th layer
(13-th layer). Conv 3 and Conv 4 are designated as targets. If not otherwise specified, we use $\epsilon=8$ as the attack power in evaluations, and use $K=5$ nearest neighbors in the kNN. We show results using $95\%$ confidence intervals over 20 (10) trials in ASK-Atk (ASK-Def) studies.

\subsection{Empirical study of ASK-Atk}\label{sec: main_atk_result}

\paragraph{Comparisons with other attacks.} We compare ASK-Atk with AdvKnn \cite{li2019advknn} and DkNN-Atk \cite{sitawarin2019robustness} on the DkNN classifier \cite{papernot2018deep}. AdvKnn and DkNN-Atk are two state-of-art attacks on kNN-based deep classifiers. We first apply different attacks on a CIFAR-10 model adversarially trained with $\epsilon=4$. We refer readers to Appendix~\ref{sec: app4} for additional details about the attack parameter settings. Performance comparisons are shown in Table~\ref{tab: attack_comp_cifar} demonstrating that ASK-Atk causes higher accuracy degradation than other attacks, for both cosine similarity and $\ell_2$ similarity measures used in the ASK loss function \eqref{eq: knn_ask}. The left figure in Figure~\ref{fig: att_defense} demonstrates that ASK-Atk outperforms other attacks for various attack power $\epsilon$. It shows almost a linear response on the ASK-Attack's accuracy degradation. The results are consistent with the observations in Figure~2 in \cite{madry2017towards}, demonstrating that increasing the attack power can cause significant accuracy degradation before 
pushing the robust accuracy to a close-to-zero region. A similar comparison is conducted on an Imagenette model adversarially trained with $\epsilon=2$. We compare ASK-Atk with other attacks using $\epsilon=4$ and $\epsilon=8$, and present the results in Table~\ref{tab: attack_comp_imagenette}. The table shows that ASK-Atk is stronger than the other attacks on Imagenette. The results on both datasets show that the attack performance of ASK-Atk does not decrease for the lower layers, unlike the other attacks. The ASK-Atk under the targeted attack and random sampling strategy are shown in Tables~\ref{tab: attack_target} and \ref{tab: attack_rand_samp} in the Appendix and demonstrate that ASK-Atk also has superior performance in these settings.

\begin{table*}[h]
\caption{{\bf ASK-Attack (ASK-Atk) outperforms AdvKnn \cite{li2019advknn} and DkNN-Atk \cite{sitawarin2019robustness} on Imagenette under $\epsilon=8$ and $\epsilon=4$.} All indicated methods are applied to the same DNN model. kNN attacks and DkNN attacks are implemented on the same layers.}
\begin{center}
\label{tab: attack_comp_imagenette}
\resizebox{1\textwidth}{!}{
\begin{tabular}{l||c|c|c|c|c|c}
\hline
\hline
& \multicolumn{3}{c|}{$\epsilon=8$}  & \multicolumn{3}{c}{$\epsilon=4$} \\ [0.5ex] 
\hline
& \multicolumn{1}{c|}{\textbf{ASK-Atk}} & \multicolumn{1}{c|}{AdvKnn}  & \multicolumn{1}{c|}{DkNN-Atk} & \multicolumn{1}{c|}{\textbf{ASK-Atk}} & \multicolumn{1}{c|}{AdvKnn}  & \multicolumn{1}{c}{DkNN-Atk}\\ [0.5ex] 
\hline
Conv 3 & \bf{34.4\%$\pm$0.24\%} & 58.8\%$\pm$0.27\% &  51.4\%$\pm$0.41\% & \bf{55.2\%$\pm$0.2\%}   & 66.2\%$\pm$0.16\% &  58.4\%$\pm$0.38\% \\
\hline
Conv 4  & \bf{26.7\%$\pm$0.17\%}  & 44.9\%$\pm$0.3\% &  56.5\%$\pm$0.35\%  & \bf{56.2\%$\pm$0.22\%}    & 68.6\%$\pm$0.25\% &  67.6\%$\pm$0.44\%\\
\hline
Conv 3,4 &   \bf{34.5\%$\pm$0.21\%}  & 50.5\%$\pm$0.26\% &  58.4\%$\pm$0.4\% &  \bf{60.2\%$\pm$0.19\%}   & 70.2\%$\pm$0.23\% &  68.1\%$\pm$0.4\%\\
\hline
\hline
\end{tabular}}
\end{center}
\end{table*}

\paragraph{Evaluations on different layer combinations.} We also show in Table~\ref{tab: attack_layers} that ASK-Atk can evaluate feature robustness on different layer combinations. ASK-Atk retains good performance when number of layers increases to five.

\begin{table*}[h]
\caption{{\bf ASK-Attack (ASK-Atk) retains good performance on different layer combinations. } All the methods are applied on the same DNN model. Experiments are conducted on CIFAR-10.}
\begin{center}
\label{tab: attack_layers}
\resizebox{0.8\textwidth}{!}{
\begin{tabular}{l||c|c|c|c|c}
\hline
\hline
& \multicolumn{1}{c|}{\begin{tabular}[c]{@{}c@{}}  Conv 4  \end{tabular}} & \multicolumn{1}{c|}{\begin{tabular}[c]{@{}c@{}}  Conv 3,4  \end{tabular}} & \multicolumn{1}{c|}{\begin{tabular}[c]{@{}c@{}}  Conv 2,3,4 \end{tabular}} & Conv 1,2,3,4 & Conv 1,2,3,4,5 \\ [0.5ex]
\hline
$K=5$ & 36.7\%$\pm$0.18\% &  41\%$\pm$0.19\% & 41.5\%$\pm$0.23\% & 41.5\%$\pm$0.21\%  & 40.1\%$\pm$0.27\% \\

\hline
$K=10$  & 38.6\%$\pm$0.14\% &  42.2\%$\pm$0.19\% & 41.7\%$\pm$0.24\%  &  41.3\%$\pm$0.2\% & 40.7\%$\pm$0.2\% \\

\hline
\hline
\end{tabular}}
\end{center}
\end{table*}

\begin{figure*}[h]
  \centering
  \includegraphics[trim=50 0 50 15,clip,width=.68\textwidth]{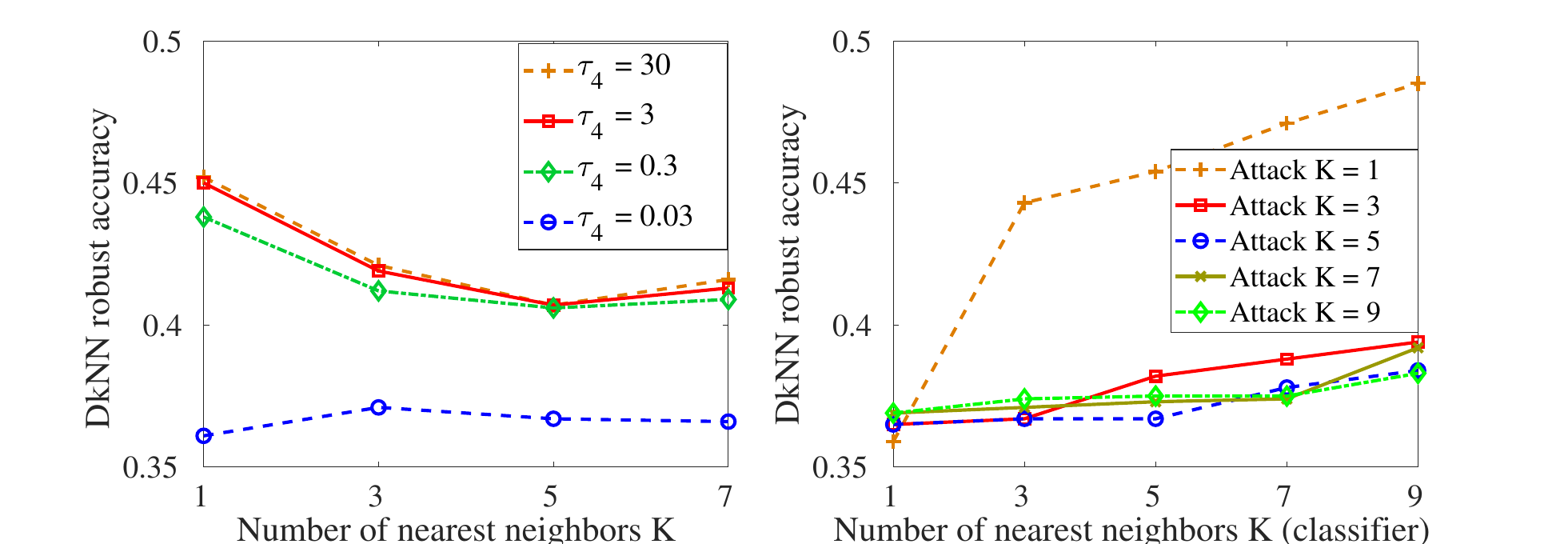}
  \caption{{\bf The ASK-Atk is robust to variations of hyperparameters $\tau_4$ and $K$}. Left: Varying $K$ has small impact on the performance of ASK-Atk when $\tau_4=0.03$. The performance of ASK-Atk remains stable even as $\tau_4$ is increased by a factor of $10,100,1000$. Right: ASK-Atk performs well when the classifier's $K$ is less or equal to the attacker's $K$, and retains relatively good performance for larger classifier $K$ when the attacker's $K$ exceeds $2$. More results can be found in Tables~\ref{tab: attack_k_large} and \ref{tab: attack_k_pos_neg} in the Appendix.}
  \label{fig: att_tau_k_vary}
\end{figure*}

\paragraph{Sensitivity to $\tau_l$.} The temperature $\tau_l$ is selected using cross-validation on a small batch, as discussed in the ASK-Attack Section. We study the sensitivity of ASK-Atk to $\tau_l$ by fixing the number of nearest neighbors $K$ to be the same in ASK-Atk and in DkNN, and study how $\tau_l$ affects the ASK-Atk performance under different values of $K$. $\tau_4=0.03$ is the optimal value on convolutional layer $4$. According to the results shown in the left figure of Figure~\ref{fig: att_tau_k_vary}, varying $K$ has only small impact on the performance of ASK-Atk when $\tau_4=0.03$. Increasing the temperature to $\tau_4$, the performance of ASK-Atk remains stable even as $\tau_4$ is increased by a factor of $10,100,1000$ times the optimal $\tau_4$. Note that the performance improves as $K$ increases when $\tau_4$ is sufficiently large. Though the performances of ASK-Atk under non-optimal $\tau_4$ are worse than the performance under optimal $\tau_4$, they are still better than AdvKnn and DkNN-Atk. 

\paragraph{Sensitivity to the nearest neighbor $K$.}
We next fix $\tau_4$ and study the gray box threat model where the attacker has no information on the value of $K$ used by DkNN. The right figure of Figure~\ref{fig: att_tau_k_vary} shows the performance of ASK-Atk using different values of $K$. ASK-Atk performs well when the classifier's $K$ is less than or equal to the attacker's $K$, and ASK-Atk performance can remain relatively good for larger values of the classifier's $K$ when the attackers value of $K$ exceeds $2$. Experimental results of larger variation of $K$ ($\geq 15$) can be found in Table~\ref{tab: attack_k_large} in the Appendix. 

\paragraph{Sensitivity to the attack implementation layers.} Here we conduct experiments on another gray-box setting where ASK-Atk is applied on different layers from the DkNN layers. We apply ASK-Atk on one of Conv 3, Conv 4, and Conv 3, 4 and also apply DkNN on the one of Conv 3, Conv 4, and Conv 3, 4. Results of these nine combinations are shown in Table~\ref{tab: attack_gray_layer}. One can see that the ASK-Atk still works well in these scenarios, although the optimal attack on a single layer is still achieved when ASK-Atk is applied on the same layer as DkNN. A good choice in the gray-box setting is to attack multi-layer, e.g., Conv 3,4 in the example shown in Table~\ref{tab: attack_gray_layer}.

\begin{table*}[h]
\caption{{\bf ASK-Atk is applied on different layers from the DkNN layers but can still cause high accuracy degradation (attack success rate). } All the methods are applied on the same DNN model. Experiments are conducted on CIFAR-10.}
\begin{center}
\label{tab: attack_gray_layer}
\resizebox{0.8\textwidth}{!}{
\begin{tabular}{l||c|c|c}
\hline
\hline
& \multicolumn{1}{c|}{\begin{tabular}[c]{@{}c@{}}  \textbf{ASK-Atk} (Conv 3)  \end{tabular}} & \multicolumn{1}{c|}{\begin{tabular}[c]{@{}c@{}}  \textbf{ASK-Atk} (Conv 4)  \end{tabular}} & \multicolumn{1}{c}{\begin{tabular}[c]{@{}c@{}}  \textbf{ASK-Atk} (Conv 3,4)  \end{tabular}} \\ [0.5ex]
\hline
Conv 3 (DkNN) & 35.1\%$\pm$0.2\% &  49\%$\pm$0.17\% & 38.6\%$\pm$0.23\%  \\

\hline
Conv 4 (DkNN)  & 55.2\%$\pm$0.25\% &  36.7\%$\pm$0.18\% & 37.1\%$\pm$0.16\%  \\
\hline
Conv 3,4 (DkNN)  &  48.2\%$\pm$0.2\% & 41.9\%$\pm$0.17\% & 41\%$\pm$0.19\%  \\

\hline
\hline
\end{tabular}}
\end{center}
\end{table*}

\paragraph{Failure cases analysis.} We compare all the attack failure cases and the success cases under the ASK-Atk from the quantitative perspective. We first calculate the prediction confidence scores of all the benign examples using 50-Nearest Neighbor. We find that the proportion of data points with a confidence score equalling one is $86.89\%$ for failure cases and $22.91\%$ for success cases. The proportion of data points with a confidence score larger than 0.95 is $93.57\%$ for failure cases and $33.88\%$ for success cases. These observations indicate that the failure cases are generally more compact with their neighbors belonging to ground truth classes and thus hard to attack. The analysis also demonstrates that ASK-Atk can be leveraged to evaluate the compactness and robustness of features extracted in hidden layers. 

\paragraph{ASK-Atk on the data space.} Although the ASK-Atk is designed to operate on inner representations of a deep neural network, we demonstrate that ASK-Atk performs well on the data space. We remark that both CIFAR-10 and Imagenette have poor kNN classification accuracy in the data space. Therefore, we tested ASK-Atk on the data space of MNIST \cite{Lecun1998gradient} and Fashion-MNIST \cite{xiao2017fashion}. The results are shown in Table~\ref{tab: attack_input}. The standard accuracy for MNIST dataset (Fashion-MNIST dataset) is $95.5\%$ ($82.3\%$) when $K=5$ and $94\%$ ($85\%$) when $K=10$. On MNIST (Fashion-MNIST), ASK-Atk can achieve $25\%$ ($40\%$) accuracy degradation under the $\ell_\infty$ norm constraint with $\epsilon=60$. 

\begin{table*}[h]
\caption{{\bf Performance of ASK-Attack (ASK-Atk) on the data space using MNIST and Fashion-MNIST.} The standard accuracy for MNIST dataset (Fashion-MNIST dataset) is $95.5\%$ ($82.3\%$) when $K=5$ and $94\%$ ($85\%$) when $K=10$.}
\begin{center}
\label{tab: attack_input}
\resizebox{0.6\textwidth}{!}{
\begin{tabular}{l||c|c}
\hline
\hline
& \multicolumn{1}{c|}{\begin{tabular}[c]{@{}c@{}}  $\epsilon=40$  \end{tabular}} & \multicolumn{1}{c}{\begin{tabular}[c]{@{}c@{}} $\epsilon=60$  \end{tabular}} \\ [0.5ex]
\hline
 MNIST ($K=5$) & 76.42\%$\pm$0.05\% &  70.5\%$\pm$0.08\%   \\

\hline
\begin{tabular}[c]{@{}c@{}}  Fashion-MNIST ($K=5$)  \end{tabular}  & 49.31\%$\pm$0.1\% &  41.2\%$\pm$0.08\%  \\

\hline
 MNIST ($K=10$)  & 77.2\%$\pm$0.1\% & 70.3\%$\pm$0.06\%  \\

\hline
\begin{tabular}[c]{@{}c@{}}  Fashion-MNIST  ($K=10$)  \end{tabular}  & 48.7\%$\pm$0.05\% & 41.8\%$\pm$0.11\% \\

\hline
\hline
\end{tabular}}
\end{center}
\end{table*}

\begin{table*}[h]
\caption{{\bf ASK-Defense (ASK-Def) outperforms Adversarial Training (AT) \cite{madry2017towards} and AT \cite{madry2017towards}+Soft Nearest Neighbor \cite{frosst2019analyzing} (AT+SNN).} The ASK-Def does a better robustification of different layers under the cosine similarity using CIFAR-10 and Imagenette. We add ASK loss to convolutional layer $4$.}
\begin{center}
\label{tab: askrt}
\resizebox{1.0\textwidth}{!}{
\begin{tabular}{l||c|c|c||c|c|c}
\hline
\hline
& \multicolumn{3}{c||}{CIFAR-10} & \multicolumn{3}{c}{Imagenette} \\ [0.5ex]
\hline
& \multicolumn{1}{c|}{\begin{tabular}[c]{@{}c@{}}\textbf{ASK-Def}\\ (\textbf{ASK loss} \\ on \textbf{Conv 4})  \end{tabular} } & \multicolumn{1}{c|}{AT} & \multicolumn{1}{c||}{\begin{tabular}[c]{@{}c@{}}AT+SNN\\ (SNN on \\ Conv 4)  \end{tabular} } & \multicolumn{1}{c|}{\begin{tabular}[c]{@{}c@{}}\textbf{ASK-Def}\\ (\textbf{ASK loss} \\ \textbf{on Conv 4})  \end{tabular}  } &  \multicolumn{1}{c|}{AT} & \multicolumn{1}{c}{\begin{tabular}[c]{@{}c@{}}AT+SNN
\\ (SNN on \\ Conv 4)  \end{tabular} } \\ [0.5ex] 
\hline
Conv 3 & \bf{42.7\%$\pm$0.16\%} & 35.1\%$\pm$0.2\% & 38.3\%$\pm$0.21\%  & \bf{37.9\%$\pm$0.19\%} & 34.4\%$\pm$0.24\% & 32.1\%$\pm$0.24\% \\
\hline
Conv 4 & \bf{43.6\%$\pm$0.19\%} & 36.7\%$\pm$0.18\% & 41.7\%$\pm$0.16\%  & \bf{30.6\%$\pm$0.17\%} & 26.7\%$\pm$0.17\% & 30.3\%$\pm$0.2\%  \\
\hline
Conv 3,4 & \bf{45.9\%$\pm$0.14\%} & 41\%$\pm$0.19\% & 41.9\%$\pm$0.18\%  & \bf{35.8\%$\pm$0.2\%} & 34.5\%$\pm$0.21\% & 34.1\%$\pm$0.21\% \\
\hline
\hline
\end{tabular}}
\end{center}
\end{table*}

\begin{table*}[h]
\caption{{\bf ASK-Defense (ASK-Def) outperforms Adversarial Training (AT) \cite{madry2017towards} and AT \cite{madry2017towards}+Soft Nearest Neighbor \cite{frosst2019analyzing} (AT+SNN) in defending AdvkNN \cite{li2019advknn} and DkNN-Atk \cite{sitawarin2019robustness}.} We use cosine similarity and CIFAR-10.}
\begin{center}
\label{tab: askrt_two}
\resizebox{1\textwidth}{!}{
\begin{tabular}{l||c|c|c||c|c|c}
\hline
\hline
& \multicolumn{3}{c||}{AdvkNN} & \multicolumn{3}{c}{DkNN-Atk} \\ [0.5ex]
\hline
& \multicolumn{1}{c|}{\begin{tabular}[c]{@{}c@{}}\textbf{ASK-Def}\\ (\textbf{ASK loss} \\ \textbf{on Conv 4})  \end{tabular} }  & \multicolumn{1}{c|}{AT} & \multicolumn{1}{c||}{\begin{tabular}[c]{@{}c@{}}AT+SNN\\ (SNN on Conv 4)  \end{tabular} } & \multicolumn{1}{c|}{\begin{tabular}[c]{@{}c@{}}\textbf{ASK-Def}\\ (\textbf{ASK loss} \\ \textbf{on Conv 4})  \end{tabular}  } &  \multicolumn{1}{c|}{AT} & \multicolumn{1}{c}{\begin{tabular}[c]{@{}c@{}}AT+SNN \\ (SNN on \\ Conv 4)  \end{tabular} } \\ [0.5ex] 
\hline
Conv 3 & \bf{61.4\%$\pm$0.24\%} & 58.9\%$\pm$0.3\% & 55.85\%$\pm$0.31\%  & \bf{62.3\%$\pm$0.26\%} & 60.7\%$\pm$0.44\% & 62\%$\pm$0.34\% \\
\hline
Conv 4 & \bf{53.4\%$\pm$0.27\%} & 50.5\%$\pm$0.34\% & 41.15\%$\pm$0.27\%  & \bf{68.1\%$\pm$0.35\%} & 50.8\%$\pm$0.39\% & 65.3\%$\pm$0.31\%  \\
\hline
Conv 3,4 & \bf{56.45\%$\pm$0.31\%} & 54.1\%$\pm$0.25\% & 45.45\%$\pm$0.17\%  & \bf{72\%$\pm$0.24\%} & 54.85\%$\pm$0.41\% & 71\%$\pm$0.28\% \\
\hline
\hline
\end{tabular}}
\end{center}
\end{table*}

\begin{table*}[ht]
\caption{{\bf ASK-Defense (ASK-Def) improves DNN prediction robustness over Adversarial Training (AT) \cite{madry2017towards} and AT \cite{madry2017towards}+Soft Nearest Neighbor \cite{frosst2019analyzing} (AT+SNN) under different PGD attack power.} Both CIFAR-10 and Imagenette are included.}
\begin{center}
\label{tab: dnn_improve}
\resizebox{0.76\textwidth}{!}{
\begin{tabular}{l||c|c|c}
\hline
\hline
CIFAR-10 &  \multicolumn{1}{c|}{$\epsilon=4$} & \multicolumn{1}{c|}{$\epsilon=6$} &  \multicolumn{1}{c}{$\epsilon=8$} \\ [0.5ex] 
\hline
\textbf{ASK-Def (Conv 4)} &  \bf{65.94\%$\pm$0.23\%}   & \bf{52.72\%$\pm$0.18\%}  & \bf{40.36\%$\pm$0.2\%}   \\

\hline
AT  & 58.34\%$\pm$0.21\%   &43.53\%$\pm$0.16\%   & 31.65\%$\pm$0.16\%   \\
\hline
AT+SNN (Conv 4)  & 59.2\%$\pm$0.18\% & 43.9\%$\pm$0.21\%   & 30.9\%$\pm$0.14\%  \\

\hline
Imagenette & \multicolumn{1}{c|}{$\epsilon=4$} & \multicolumn{1}{c|}{$\epsilon=6$} & \multicolumn{1}{c}{$\epsilon=8$} \\ [0.5ex]
\hline
\textbf{ASK-Def (Conv 4)}  & \bf{55.2\%$\pm$0.25\%}  & \bf{40.2\%$\pm$0.15\%}  &  \bf{29.9\%$\pm$0.13\%}  \\
\hline
AT & 49.1\%$\pm$0.28\%  & 32.3\%$\pm$0.22\%  & 19.6\%$\pm$0.1\%  \\
\hline
AT+SNN (Conv 4)  & 50.5\%$\pm$0.19\%  &34.9\%$\pm$0.15\%  & 23.1\%$\pm$0.07\%  \\
\hline
\hline
\end{tabular}}
\end{center}
\end{table*}

\subsection{Empirical study of ASK-Def}\label{sec: results_askrt}

\paragraph{Improving DkNN robustness by ASK-Def.} We first implement the ASK-Atk as the evaluation method to test DkNN robustness on models trained by different robust training methods. 
To demonstrate the effectiveness of ASK-Def, we compare it to \textbf{(i)} using Adversarial Training (AT) \cite{madry2017towards}; and \textbf{(ii)} AT \cite{madry2017towards}+ Soft Nearest Neighbor (SNN) \cite{frosst2019analyzing}, where SNN was proposed as a regularizer to improve feature representation and model generalization. The results are shown in Tables~\ref{tab: askrt}, which demonstrates that there is additional robustness brought on by the auxiliary ASK loss term. We empirically find that applying ASK loss on the Conv 4 is enough to robustify both Conv 3 and Conv 4. One can see that the model trained by ASK-Def attains the highest robust accuracy against ASK-Atk on the two datasets across on different layers. Specifically, on CIFAR-10/Imagenette, ASK-Def improves the robustness of DkNN over AT (AT+SNN) by $\geq 6.9\%/3.5\%$ ($\geq 1.9\%/0.3\%$) on single selected layer, and $4.8\%/1.3\%$ ($4\%/1.7\%$) on combined layers. The right panel in Figure~\ref{fig: att_defense} demonstrates that the model trained by ASK-Def has improved DkNN robustness over that of the model trained by AT over a range of attack power $\epsilon$. We also demonstrate that the model trained by ASK-Def improves robustness over a range of values of $K$, see Figure~\ref{fig: k_vary_rt} in the Appendix. Our theoretical analysis indicates that ASK-Def can improve universal robustness beyond ASK-Atk. We next conduct experiments on AdvKnn and DkNN-Atk. Table~\ref{tab: askrt_two} shows that ASK-Def can defend against various attacks and outperforms AT and AT+SNN. Specifically, ASK-Def outperforms the AT+SSN (AT) by $12\%$ ($17\%$) under the AdvkNN (DkNN-Atk) on the convolutional layer 4. These results are comparable to the improvements shown in Table~\ref{tab: askrt}, and we believe the results in Table~\ref{tab: askrt_two} are strong enough to show the additional robustness brought on by the auxiliary ASK loss term. ASK-Def can also be combined with variants of adversarial training, e.g., TRADES \cite{zhang2019theoretically}. Additional experimental results on comparing with TRADES are shown in Table~\ref{tab: def_trades} in the Appendix. We choose TRADES because AT and TRADES are the two most widely used robust training methods. Although ASK-Def can be combined with other variants of AT, we believe that current results are enough to show the effectiveness of our defense method.

\paragraph{Improving DNN robustness.} As illustrated in Section~\ref{sec: RT}, ASK-Def is able to improve the robustness of both DkNN and DNN at the same time. For the PGD attack, Table~\ref{tab: dnn_improve} shows that ASK-Def can improve DNN robustness under different attack power levels, without appreciable loss of standard accuracy. Compared the $\epsilon=8$ column in Table~\ref{tab: dnn_improve} with the results shown in Table~\ref{tab: askrt} on Conv 4, one can see that the differences of robust accuracy values on CIFAR-10/Imagenette for ASK-Def are $3.24\%/0.7\%$, while the differences of robust accuracy values for AT are $5.05\%/7.1\%$, indicating that the robust accuracy gap between DkNN and DNN becomes smaller in the model trained by ASK-Def.

\section{Conclusion}
We presented a novel attack and defense strategy for kNN-based deep classifiers. These strategies both include our proposed Adversarial Soft kNN (ASK) loss term, which we demonstrated is a lower bound on the mutual information between perturbed inputs and the in-class reference data. Maximization of the ASK loss led to  our proposed ASK-Attack (ASK-Atk). Experiments show that ASK-Atk outperforms previous kNN-based attacks. A minimax optimization strategy applied to the ASK loss was introduced leading to our robust training method called ASK-Defense (ASK-Def). Experiments demonstrate that ASK-Def provides additional robustness for kNN-based deep classifiers compared to conventional adversarial training. ASK-Atk and ASK-Def are applicable to a wide array of non-parametric metric learning applications beyond the computer vision applications and kNN-DNN hybrid system developed here.

\appendices

\section{Proof of Proposition 1}\label{sec: app1}

Our proof follows the similar proof technique in \cite{oord2018representation}. We will show that (i) the optimal similarity measurement $\left [S(f_{\theta_{l}}; \mathbf x^\prime, X_c^{l}) \right ]_{\rm opt}$ is proportional to density ratio between the conditional distribution $p(X_c^{l} | \mathbf x^\prime)$ and the marginal distribution $p(X_c^{l})$, where $X_c^{l}$ denotes the reference data from class $c \in [C]$  (ii) Minimizing the ASK loss maximizes a lower bound on the mutual information between the perturbed input and the in-class-reference data. Given a reference data set $X^{l}=\{ X^{l+},X_c^{l-}\}, c \in [C]/y$ containing one in-class reference sample from the distribution $p(\cdot|\mathbf x^\prime)$ and $C-1$ out-of-class reference samples from the distribution $p(\cdot)$, we optimize the ASK loss to correctly select the in-class reference sample out of the given set. In other words, the optimal probability for the ASK loss should imply the fact that $X^{l+}$ comes from $p(\cdot|\mathbf x^\prime)$ and $X_c^{l-}, c \in [C]/y$ come from $p(\cdot)$. We use $P(X_c^l = l+ | X^{l}, \mathbf x_i^\prime)$ to represent such probability. We then have Lemma~\ref{lemma1} for the optimal choice of the similarity measurement $S(f_{\theta_{l}}; \mathbf x^\prime, X_c^{l})$ under the optimal probability by minimizing the ASK loss.
\begin{lemma}\label{lemma1}
$\left [S(f_{\theta_{l}}; \mathbf x^\prime, X_c^{l}) \right ]_{\rm opt} \propto \frac{P(X_c^{l} | \mathbf x^\prime)}{P(X_c^{l})}$.
\end{lemma}

The proof details are shown as follows.
\begin{proof}
$\frac{ S(f_{\theta_{l}}; \mathbf x^\prime, X_c^{l}) 
    }{
    S(f_{\theta_{l}}; \mathbf x^\prime, X_c^{l}) + 
   \sum_{i=1,i\neq c}^C S(f_{\theta_{l}}; \mathbf x^\prime, X_i^{l})  
    }$ can be viewed as the prediction score of predicting $X_c^{l}$ to belonging to the distribution  $p(\cdot|\mathbf x^\prime)$. First, we calculate the optimal probability of $P(X_c^{l} = l+ | X^{l}, \mathbf x^\prime)$. We have
\begin{align}\label{eq: prop_opt}
& P(X_c^{l} = l+ | X^{l}, \mathbf x^\prime)\nonumber \\&= \frac{p(X_c^{l}|\mathbf x^\prime)\prod_{i \not= c}p(X_{i}^{l})}{p(X_c^{l}|\mathbf x^\prime)\prod_{i \not= c}p(X_{i}^{l}) + \sum_{i \not= c}p(X_{i}^{l}|\mathbf x^\prime)\prod_{j \not= i}p(X_{j}^{l})}\nonumber \\&= \frac{\frac{p(X_c^{l}|\mathbf x^\prime)}{p(X_c^{l})}}{\frac{p(X_c^{l}|\mathbf x^\prime)}{p(X_c^{l})}+\sum_{i \not= c} \frac{p(X_{i}^{l}|\mathbf x^\prime)}{p(X_{i}^{l})}},
\end{align}

Compare the result of \eqref{eq: prop_opt} with $\frac{ S(f_{\theta_{l}}; \mathbf x^\prime, X_c^{l}) 
    }{
    S(f_{\theta_{l}}; \mathbf x^\prime, X_c^{l}) + 
   \sum_{i\neq c} S(f_{\theta_{l}}; \mathbf x^\prime, X_{i}^{l})  
    }$, the proof is done.
\end{proof}

Leveraging Lemma \ref{lemma1}, we provide the proof of Proposition~\ref{eq: thm} below.

\begin{proof}
\begin{align}
& L_{\rm ask}^l \geq L_{\rm ask}^l(OPT) \label{eq:1} \\&=- \mathbb E_{(X^{l}, \mathbf x^\prime)} \nonumber \\& \log \left [ \frac{ S(f_{\theta_{l}}; \mathbf x^\prime, X^{l+}) 
    }{
    S(f_{\theta_{l}}; \mathbf x^\prime, X^{l+}) + 
   \sum_{c=1,c\neq y}^C S(f_{\theta_{l}}; \mathbf x^\prime, X_{c}^{l-})  
    } \right ]_{opt} \label{eq:2} \\&= - \mathbb E_{(X^{l}, \mathbf x^\prime)} \log\frac{\frac{p(X^{l+}|\mathbf x^\prime)}{p(X^{l+})}}{\frac{p(X^{l+}|\mathbf x^\prime)}{p(X^{l+})}+\sum_{c \neq y} \frac{p(X_{c}^{l-}|\mathbf x^\prime)}{p(X_{c}^{l-})}} \label{eq:3} \\
    &= \mathbb E_{(X^{l}, \mathbf x^\prime)} \log \left [  1 + \frac{p(X^{l+})}{p(X^{l+}|\mathbf x^\prime)} \sum_{c \neq y} \frac{p(X_{c}^{l-}|\mathbf x^\prime)}{p(X_{c}^{l-})}  \right ] \label{eq:4} \\
    & \approx \mathbb E_{(X^{l}, \mathbf x^\prime)} \log \left [  1 + \frac{p(X^{l+})}{p(X^{l+}|\mathbf x^\prime)} (C-1) \mathbb E_{X^{l-}} \frac{p(X_{c}^{l-}|\mathbf x^\prime)}{p(X_{c}^{l-})}  \right ] \label{eq:5} \\
    & = \mathbb E_{ (X^{l+}, \mathbf x^\prime) } \log \left [  1 + \frac{p(X^{l+})}{p(X^{l+}|\mathbf x^\prime)} (C-1) \right ] \label{eq:6} \\
    & \geq  \mathbb E_{ (X^{l+}, \mathbf x^\prime) } \log  (C-1)\frac{p(X^{l+})}{p(X^{l+}|\mathbf x^\prime)} \label{eq:7} \\
    & = \mathbb E_{(X^{l+}, \mathbf x^\prime)} \log  \frac{p(X^{l+})p(\mathbf x^\prime)}{p(X^{l+}, \mathbf x^\prime)} + \log (C-1) \label{eq:8} \\
    & = - \mathrm{MI} (X^{l+},  \mathbf x^\prime) + \log (C-1),
\end{align}
\end{proof}
\noindent where \eqref{eq:1} holds because any loss value is larger or equal to the optimal value. \eqref{eq:3}  comes from Lemma \ref{lemma1}. For simplicity, we only consider one data point from each class. $(C-1)$ will be replace by $K(C-1)$ if each class has $K$ reference data. \eqref{eq:5} becomes more accurate when the number of classes or number of reference data in each class increases. In practice, the number of classes in most classification problems is large enough for the approximation to work decently well. 
One can also check that the ASK losses under two forms of $S(\cdot, \cdot)$ (used for design of the attack and the defense) lower bound the mutual information. They can be treated equivalent under Proposition~\ref{eq: thm}.

\section{Algorithm for ASK-Atk}\label{sec: app2}
We formulate ASK-Atk as an optimization problem that maximizes the ASK loss under the given constraints on perturbations and adversarial examples. Algorithm \ref{alg: ask_atk} shows the details of implementing ASK-Atk, which includes steps of gradient ascent and projections. 
\begin{algorithm}
\caption{ASK-Atk}
   \label{alg: ask_atk}
\begin{algorithmic}
\REQUIRE Model weights before the target layer $l$, $\theta_l$; attack steps $s$; step size $\kappa$; attack power $\epsilon$; reference data points $X^{l}$.
\STATE Uniformly initialize $\mathbf \delta$ from $[-\epsilon, \epsilon]$, and $\mathbf x^{\prime}[0] = \mathbf x+\mathbf \delta$.
\FOR{$j = 1, \cdots, s$}
\STATE $\mathbf x^{\prime}[j] \leftarrow  \mathbf x^{\prime}[j-1] + \kappa {\rm{sign}}\big(\nabla_{\mathbf x^{\prime}} \sum_{l \in \mathcal{L}} \omega_l L_{\mathrm{ask}}^l(\theta_{l}, \mathbf x^{\prime}[j-1], X^{l})\big)$.
\STATE $\mathbf x^{\prime}[j] \leftarrow {\rm Clip}_{[0,1]} \big( \mathbf x^{\prime}[j] \big) $.
\STATE $\mathbf x^{\prime}[j] \leftarrow \mathbf x^{\prime}[j] + {\rm Clip}_{[-\epsilon, \epsilon]} ( \mathbf \delta )$.
\ENDFOR

\STATE {\bfseries Output:} $\mathbf x$.
\end{algorithmic}
\end{algorithm}

\section{Algorithm for ASK-Def}\label{sec: app3}

ASK-Def is constructed in two stages corresponding to adversarial examples generation and model update. The inner maximization can be efficiently solved using gradient ascent with projection, and the model weights are updated using gradient descent. The pseudocode for the full ASK-Def procedure is given in Algorithm~\ref{alg: ASK-Def}. $\rm{Attack}^{\rm{CE}}$ and $\rm{Attack}^{\rm{ASK}}$ refer to our generators on DNN cross-entropy and on ASK loss. $\rm{Update}^{\rm{CE}}$ and $\rm{Update}^{\rm{ASK}}$ denote the gradient updates of first and second terms in model update stage.

\begin{algorithm}[H]
\caption{ASK-Defense}
   \label{alg: ASK-Def}
\begin{algorithmic}
\REQUIRE Initial model $\theta$; model weights before the target layer $l$, $\theta_l$; number of total training epochs $E$; number of batches in one epoch $T$; attack power $\epsilon$; penalty parameter $\lambda$; number of reference data points in each class $K$; reference data $S_{rt}$ in each batch.
\FOR{$e = 1, \cdots, E$}
\FOR{$t = 1, \cdots, T$}
\STATE $S_t(\epsilon) \longleftarrow {\rm{Attack}}^{{\rm{CE}}}(\theta, S_t)$. $S_t^l(\epsilon) \longleftarrow {\rm{Attack}}^{{\rm{ASK}}}(\theta_l, S_t)$.
\STATE $\theta \longleftarrow {\rm{Update}}^{{\rm{CE}}}(\theta, S_t(\epsilon)) + \lambda {\rm{Update}}^{{\rm{ASK}}}(\theta_l, S_t^l(\epsilon), S_{rt}(\epsilon))$.
\ENDFOR
\ENDFOR

\STATE {\bfseries Output:} $\theta$.
\end{algorithmic}
\end{algorithm}

\section{Datasets, models, and settings}\label{sec: app4}

\paragraph{Datasets.} We mainly utilize two datasets, CIFAR-10 and Imagenette (10 sub-classes of ImageNet), to demonstrate the effectiveness of our methods. CIFAR-10 contains $50000$ training samples and $10000$ test samples. Each image is in the size of $32\times 32 \times 3$. Imagenette contains 10 classes from ImageNet and includes $9469$ training samples and $3925$ test samples. We rescale each image to $128 \times 128 \times 3$. These two datasets are also used in many state-of-the-art empirical attacks and defenses. As a comparison, DkNN and DkNN-Atk only consider simpler datasets like MNIST and SVHN. In the experiments on data space, we utilize MNIST \cite{Lecun1998gradient} and Fashion-MNIST \cite{xiao2017fashion}. The reason is that both CIFAR-10 and Imagenette have poor kNN classification accuracy in the data space. Both MNIST and Fashion-MNIST consist of $60000$ training samples and $10000$ test samples.

\paragraph{Models.} VGG16 and ResNet18 architectures are used for CIFAR-10 and Imagenette, respectively. The CIFAR-10 model used for attack evaluation (the same baseline model used in ASK-Def comparison) is adversarially trained with $\epsilon=4$. The Imagenette model used for attack evaluation (the same baseline model used in ASK-Def comparison) is adversarially trained with $\epsilon=2$. We apply Deep k-Nearest Neighbor (DkNN) \cite{papernot2018deep} on hidden layers of VGG16 and ResNet18.

\paragraph{DkNN layer selection.} Note that both VGG16 and ResNet18 contain \textbf{five} convolutional layer blocks. We name the $i$th block Conv $i$. Conv 5 is close to the output layer, thus the robustness is similar to the output, which is lower than Conv 4 and Conv 3. In contrast, Conv 1 and Conv 2 are shallow layers that have not learned good features and have low standard/robust accuracy. We consider the third and fourth convolutional layer blocks of VGG16 and ResNet18 since these two layers have the highest robust accuracy compared to others. We also study different combinations of layers and show results in Table~\ref{tab: attack_layers}.

\paragraph{Total amount of compute and type of resources.} We use 1 GPU (Tesla V100) with 64GB memory and 2 cores for all the experiments.

\paragraph{Attack settings.} We compare ASK-Atk with AdvKnn \cite{li2019advknn} and DkNN-Atk \cite{sitawarin2019robustness} on DkNN classifiers. We use $20$ iteration steps for all the attacks. For AdvKnn, we train the deep kNN block with $100$ epochs. For DkNN-Atk, we select the optimal hyperparameter in the sigmoid function on each attack. We consider the third and fourth convolutional layers of VGG16 and ResNet18 since these two layers have the highest robust accuracy compared to others. For all the tests on CIFAR-10 (Imagenette), we randomly select $20000$ ($8500$) training samples that are equally distributed across different classes. All the test data examples are used for evaluation. We show results using $95\%$ confidence intervals over 20 trials. If not otherwise specified, we use $\epsilon=8$ for evaluating both CIFAR-10 and Imagenette and use the nearest neighbor number $K=5$ by default. We find the optimal $\tau_l$ (and $\omega_l$ for combined layers) in the first batch and then fix these parameters in the following batches. For both the CIFAR-10 and Imagenette baseline models, we use $\tau_3=\tau_4=0.03$. For the combined layers of the CIFAR-10 baseline model, we use $\omega_3=\omega_4=0.5$. For the combined layers of the Imagenette baseline model, we use $\omega_3=0.1$ and $\omega_4=0.9$.

\begin{table*}[h]
\caption{{\bf The targeted attack of ASK-Attack (ASK-Atk) can achieve the similarly good performance as non-targeted attack of ASK-Atk on both CIFAR-10 and Imagenette} All the methods are applied on the same DNN model. ASK-Atk and DkNN are implemented on the same layers and use the $\ell_2$ similarity in the ASK loss \eqref{eq: knn_ask}.}
\begin{center}
\label{tab: attack_target}
\resizebox{0.82\textwidth}{!}{
\begin{tabular}{l||c|c|c|c}
\hline
\hline
& \multicolumn{2}{c|}{CIFAR-10} & \multicolumn{2}{c}{Imagenette} \\ [0.5ex] 
\hline
& \multicolumn{1}{c|}{\begin{tabular}[c]{@{}c@{}}  \textbf{ASK-Atk} \\ (non-targeted)  \end{tabular}} & \multicolumn{1}{c|}{\begin{tabular}[c]{@{}c@{}}  \textbf{ASK-Atk} \\ (targeted)  \end{tabular}} & \multicolumn{1}{c|}{\begin{tabular}[c]{@{}c@{}}  \textbf{ASK-Atk} \\ (non-targeted)  \end{tabular}}  & \multicolumn{1}{c}{\begin{tabular}[c]{@{}c@{}}  \textbf{ASK-Atk} \\ (targeted)  \end{tabular}} \\ [0.5ex] 
\hline
Conv 3 & 32.21\%$\pm$0.16\% &  32.4\%$\pm$0.13\% & 35.9\%$\pm$0.18\% &  29.4\%$\pm$0.2\% \\

\hline
Conv 4  & 34.46\%$\pm$0.16\% &  35.9\%$\pm$0.14\% & 26.6\%$\pm$0.13\% &  27.4\%$\pm$0.11\% \\

\hline
Conv 3,4  &  39.26\%$\pm$0.17\% & 39.9\%$\pm$0.14\% & 35.2\%$\pm$0.2\% &  33.9\%$\pm$0.22\% \\

\hline
\hline
\end{tabular}}
\end{center}
\end{table*}

\begin{table*}[h]
\caption{{\bf The random sampling strategy of ASK-Attack (ASK-Atk) performs not as good as using the exact kNN search, but still outperforms AdvKnn \cite{li2019advknn} and DkNN-Atk \cite{sitawarin2019robustness} on both CIFAR-10 and Imagenette} All the methods are applied on the same DNN model. ASK-Atk and DkNN are implemented on the same layers and use the cosine similarity in the ASK loss \eqref{eq: knn_ask}.}
\begin{center}
\label{tab: attack_rand_samp}
\resizebox{0.87\textwidth}{!}{
\begin{tabular}{l||c|c|c|c}
\hline
\hline
& \multicolumn{2}{c|}{CIFAR-10} & \multicolumn{2}{c}{Imagenette} \\ [0.5ex] 
\hline
& \multicolumn{1}{c|}{\textbf{ASK-Atk}} & \multicolumn{1}{c|}{\begin{tabular}[c]{@{}c@{}}  \textbf{ASK-Atk} \\ (random sampling)  \end{tabular}} & \multicolumn{1}{c|}{\textbf{ASK-Atk}}  & \multicolumn{1}{c}{\begin{tabular}[c]{@{}c@{}}  \textbf{ASK-Atk} \\ (random sampling)  \end{tabular}} \\ [0.5ex] 
\hline
Conv 3 & 35.1\%$\pm$0.2\% &  46.4\%$\pm$0.35\% & 34.4\%$\pm$0.24\% &  47.5\%$\pm$0.38\% \\

\hline
Conv 4  & 36.7\%$\pm$0.18\% &  39.2\%$\pm$0.24\% & 26.7\%$\pm$0.17\% &  30.8\%$\pm$0.21\% \\

\hline
Conv 3,4  &  41\%$\pm$0.19\% & 49.4\%$\pm$0.3\% & 34.5\%$\pm$0.21\% &  49.1\%$\pm$0.34\% \\

\hline
\hline
\end{tabular}}
\end{center}
\end{table*}

\paragraph{Defense settings.}  Our main goal is to harden both Conv 3 and Conv 4 via ASK-Def. Note that Conv 4 will not be robustified if applying ASK loss to Conv 3. We find that applying ASK loss to Conv 4 can reach similar robustness compare to applying ASK loss to both Conv 3 and Conv 4, and in the meantime, enjoys less computational cost. In ASK-Def, we set $\lambda=1$, and $\tau_4=0.1$. We show results using $95\%$ confidence intervals over 10 trials. Throughout the experiments on defense, we implement ASK-Atk, DkNN-Atk, AdvKnn as the evaluation methods to test DkNN robustness on models trained by Adversarial Training (AT) \cite{madry2017towards}, AT \cite{madry2017towards}+ Soft Nearest Neighbor (SNN) \cite{frosst2019analyzing}, and the models trained by the ASK-Def. Similarly, we apply SSN to Conv 4. When evaluating the models trained by different robust training methods, we find the optimal $\tau_l$ and $\omega_l$ in the same way as discussed in the attack setting.

\section{Additional experimental results of ASK-Atk}\label{sec: app5}

\paragraph{Targeted attack and random sampling strategy.}
For the targeted attack we simply replace $\sum_{c\neq y}^C S(f_{\theta_{l}}; \mathbf x + \mathbf \delta, X_{c}^{l-})$ with $S(f_{\theta_{l}}; \mathbf x + \mathbf \delta, X_{c_t}^{l-})$ in the ASK loss. Here we select the target class by calculating the average distance to $\mathbf x$. We show that the targeted attack of ASK-Atk can achieve the similarly good performance as non-targeted attack on both CIFAR-10 and Imagenette, as shown in Table~\ref{tab: attack_target}. In Table~\ref{tab: attack_rand_samp}, we show how the attack perform when we select $X_{y}^{l-}$ randomly. The results indicate that selecting $X_{y}^{l-}$ randomly can still provide a stronger attack than AdvKnn \cite{li2019advknn} and DkNN-Atk \cite{sitawarin2019robustness} (results for these attacks are shown in Tables~\ref{tab: attack_comp_cifar}, \ref{tab: attack_comp_imagenette}). To achieve an optimal attack, $X_{y}^{l-}$ should be selected using kNN search.

\paragraph{Experiments when varying $K$ (Attack) and $K$ (Classifier) from 15 to 75.} Here we focus on larger $K$ ($K \geq 9$).The following Table~\ref{tab: attack_k_large} shows that the ASK-Atk can achieve a good performance when $K$ increases from 15 to 75.

\begin{table*}[h]
\caption{{\bf ASK-Attack (ASK-Atk) retains good performance when varying $K$ (Attack) and $K$ (Classifier) from 15 to 75. } Each entry shows the classification accuracy of the classifier. ASK-Atk and DkNN are applied on Conv 3, 4.}
\begin{center}
\label{tab: attack_k_large}
\resizebox{0.9\textwidth}{!}{
\begin{tabular}{l||c|c|c|c}
\hline
\hline
& \multicolumn{1}{c|}{$K=15$ (Classifier)} & \multicolumn{1}{c|}{$K=35$ (Classifier)} & \multicolumn{1}{c|}{$K=55$ (Classifier)} & \multicolumn{1}{c}{$K=75$ (Classifier)} \\ [0.5ex]
\hline
\begin{tabular}[c]{@{}c@{}}  $K=15$ \\ (Attack)  \end{tabular}  & 40.72\%$\pm$0.16\% &  43.32\%$\pm$0.2\% & 45.7\%$\pm$0.23\% & 46.8\%$\pm$0.21\% \\

\hline
\begin{tabular}[c]{@{}c@{}}  $K=35$ \\ (Attack)  \end{tabular}  & 40.6\%$\pm$0.19\% & 40.2\%$\pm$0.18\% & 41.63\%$\pm$0.16\% & 43.2\%$\pm$0.11\% \\

\hline
\begin{tabular}[c]{@{}c@{}}  $K=55$ \\ (Attack)  \end{tabular}  &  40.2\%$\pm$0.22\% & 40.4\%$\pm$0.15\% & 40.2\%$\pm$0.21\% & 41.33\%$\pm$0.13\% \\
\hline
\begin{tabular}[c]{@{}c@{}}  $K=75$ \\ (Attack)  \end{tabular}  & 41.57\%$\pm$0.14\% & 40.9\%$\pm$0.16\% & 41.11\%$\pm$0.14\% & 40.8\%$\pm$0.25\% \\

\hline
\hline
\end{tabular}}
\end{center}
\end{table*}

\paragraph{When $K$ is different for $X_{y}^{l+}$ and $X_{y}^{l-}$.} Here we study the performance of ASK-Atk when $K$ is different for  $X_{y}^{l+}$ and $X_{y}^{l-}$. We use $K^+$ and $K^-$ to represent the $K$ for $X_{y}^{l+}$ and $X_{y}^{l-}$, respectively. We vary $K^+$ from $3$ to $7$, and vary $K^-$ from $1$ to $9$. The number of nearest neighbors in DkNN is always set as $K^+$. We conduct all experiments on CIFAR-10 to Conv 4. We use the default $\epsilon=8$. The results are shown in Figure~\ref{tab: attack_k_pos_neg}. One can see that ASK-Atk can still achieve good performances when $X_{y}^{l+}$ and $X_{y}^{l-}$ have different $K$. When $K^+ \geq 3$, ASK-Atk becomes stronger for $K^- \geq 3$.

\begin{table*}[h]
\caption{{\bf ASK-Atk performs well when $K$ is different for $X_{i}^{l+}$ and $X_{i}^{l-}$. } Here we use $K^+$ and $K^-$ to represent the $K$ for $X_{i}^{l+}$ and $X_{i}^{l-}$, respectively. All experiments are conducted on CIFAR-10 to convolutional layer $4$.}
\begin{center}
\label{tab: attack_k_pos_neg}
\resizebox{0.95\textwidth}{!}{
\begin{tabular}{l||c|c|c|c|c}
\hline
\hline
& \multicolumn{1}{c|}{$K^-=1$} & \multicolumn{1}{c|}{$K^-=3$} & \multicolumn{1}{c|}{$K^-=5$} & \multicolumn{1}{c|}{$K^-=7$} & \multicolumn{1}{c}{$K^-=9$} \\ [0.5ex]
\hline
$K^+=3$ & 40.8\%$\pm$0.2\% &  36\%$\pm$0.14\% & 36.1\%$\pm$0.19\% &  36.3\%$\pm$0.17\% & 36.4\%$\pm$0.22\% \\

\hline
$K^+=5$  & 42.3\%$\pm$0.23\% &  36.9\%$\pm$0.18\% & 36.7\%$\pm$0.18\% &  35.9\%$\pm$0.12\% & 35.9\%$\pm$0.13\% \\

\hline
$K^+=7$  &  42.8\%$\pm$0.18\% & 38\%$\pm$0.15\% & 36.6\%$\pm$0.2\% &  36.3\%$\pm$0.13\% & 36\%$\pm$0.17\% \\

\hline
\hline
\end{tabular}}
\end{center}
\end{table*}

\paragraph{Additional notes on our results.} Despite the higher success rate of ASK-Atk, we also find that the ASK-Atk is much faster than DkNN-Atk and AdvKnn. Under our setting, ASK-Atk is around $20$ times faster.

\begin{figure}[h]
  \centering
  \includegraphics[trim=0 0 0 0,clip,width=.48\textwidth]{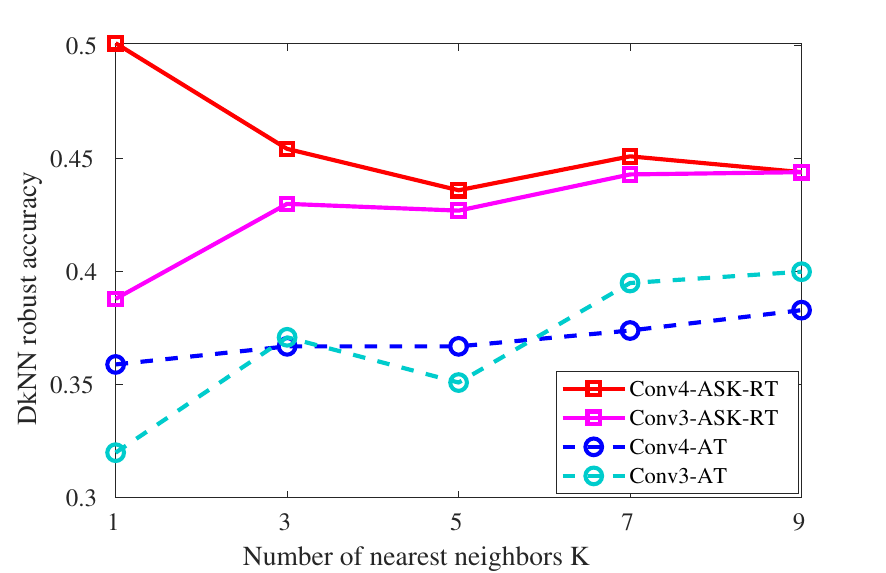}
  \caption{{\bf ASK-Defense (ASK-Def) outperforms Adversarial Training (AT) \cite{madry2017towards} when attacks and DkNN vary the number of nearest neighbors $K$}. The ASK-Def does a better robustification of different layers using CIFAR-10. We add ASK loss to convolutional layer $4$ and test on convolutional layers 3 and 4.}
  \label{fig: k_vary_rt}
\end{figure}

\section{Additional experimental results of ASK-Def}\label{sec: app6}

\paragraph{Robustness under different $K$.} We also check how the model (trained by ASK-Def) robustness changes when $K$ in ASK-Atk and DkNN varies. One can see from Figure~\ref{fig: k_vary_rt} that the model trained by ASK-Def achieves higher robustness under different $K$ compared to the model trained by conventional adversarial training \cite{madry2017towards}.

\begin{table*}[h]
\caption{{\bf ASK-Defense (TRADES) outperforms TRADES \cite{zhang2019theoretically}. } We combine ASK-Defense with the TRADES loss term. The ASK-Defense (TRADES) has a better performance on the robustification of different layers under the cosine similarity using CIFAR-10. We evaluate the robustness of different defenses using ASK-Atk with $\epsilon=4, 8$.}
\begin{center}
\label{tab: def_trades}
\resizebox{0.95\textwidth}{!}{
\begin{tabular}{l||c|c|c|c}
\hline
\hline
& \multicolumn{1}{c|}{\begin{tabular}[c]{@{}c@{}}  \textbf{ASK-Defense (TRADES)} \\ ($\epsilon=4$)  \end{tabular}} & \multicolumn{1}{c|}{\begin{tabular}[c]{@{}c@{}}  TRADES   \\ ($\epsilon=4$)  \end{tabular}} & \multicolumn{1}{c|}{\begin{tabular}[c]{@{}c@{}}  \textbf{ASK-Defense (TRADES)} \\ ($\epsilon=8$)  \end{tabular}} & \multicolumn{1}{c}{\begin{tabular}[c]{@{}c@{}}  TRADES \\ ($\epsilon=8$)  \end{tabular}} \\ [0.5ex]
\hline
Conv 3 & \bf{64.63\%$\pm$0.25\%} &  64.11\%$\pm$0.2\% & \bf{38.3\%$\pm$0.12\%} & 37.56\%$\pm$0.1\% \\

\hline
Conv 4  & \bf{67.58\%$\pm$0.22\%} &  64.74\%$\pm$0.16\% & \bf{36.92\%$\pm$0.1\%} & 30.62\%$\pm$0.07\%  \\
\hline
Conv 3,4  &  \bf{67.79\%$\pm$0.19\%} & 65.26\%$\pm$0.28\% & \bf{40.05\%$\pm$0.17\%} & 33.08\%$\pm$0.07\% \\

\hline
\hline
\end{tabular}}
\end{center}
\end{table*}

\begin{table*}[h]
\caption{{\bf ASK-Defense (ASK-Def) outperforms Adversarial Training (AT) \cite{madry2017towards} and AT \cite{madry2017towards}+Soft Nearest Neighbor \cite{frosst2019analyzing} (AT+SNN) under the ASK-Attack with an increasing number of attack iteration steps. } We fix the attack power $\epsilon=8$ and use the cosine similarity. All experiments are conducted on CIFAR-10 and are applied on convolutional layer 4.  Other settings follow the default settings in the paper.}
\begin{center}
\label{tab: def_iter}
\resizebox{0.95\textwidth}{!}{
\begin{tabular}{l||c|c|c|c|c}
\hline
\hline
& \multicolumn{1}{c|}{$iter=20$} & \multicolumn{1}{c|}{$iter=40$} & \multicolumn{1}{c|}{$iter=60$} & \multicolumn{1}{c}{$iter=80$} & \multicolumn{1}{c}{$iter=100$} \\ [0.5ex]
\hline
\textbf{ASK-Def} & 43.6\%$\pm$0.19\% & 43.1\%$\pm$0.19\% & 43.02\%$\pm$0.12\% & 42.95\%$\pm$0.18\% & 42.87\%$\pm$0.16\% \\

\hline
AT  & 36.7\%$\pm$0.18\% &  36.21\%$\pm$0.15\% & 36.15\%$\pm$0.11\% & 35.98\%$\pm$0.13\% & 35.96\%$\pm$0.1\%  \\
\hline
AT+SNN  &  41.7\%$\pm$0.16\% & 41.04\%$\pm$0.15\% & 40.93\%$\pm$0.2\% & 40.9\%$\pm$0.18\% & 40.9\%$\pm$0.15\% \\

\hline
\hline
\end{tabular}}
\end{center}
\end{table*}

\paragraph{Variants of ASK-Defense.} ASK-Defense can be combined with variants of adversarial training, e.g., TRADES \cite{zhang2019theoretically} or other state-of-the-art robust training methods. Here we replace the first loss term (together with the adversarial generator) with TRADES loss term. We name the ASK-Defense with TRADES loss term to ASK-Defense (TRADES). We conduct additional experiments on comparing with TRADES. Table~\ref{tab: def_trades} shows that ASK-Defense (TRADES) outperforms TRADES.

\paragraph{Defense against attacks with large iteration number.} We conduct experiments on increasing the iteration steps of ASK-Attack. Experimental results in Table~\ref{tab: def_iter} show that the ASK-Defense (ASK-Def) outperforms Adversarial Training (AT) \cite{madry2017towards} and AT \cite{madry2017towards}+Soft Nearest Neighbor \cite{frosst2019analyzing} (AT+SNN) under the ASK-Attack with an increasing number of attack iteration steps. Another observation is that increasing the number of iteration steps has little impact on the performance when  $iter \geq 20$.

\section{Failure Cases Analysis and Social Impact}

\paragraph{Failure cases analysis.} Here we provide more analysis on those failure cases. A consistent observation with the analysis in the main paper is that the average similarity gap between the largest class and the runner-up class for failure cases is 0.1555, while it is only 0.0519 for success cases (here we use cosine similarity). Again, these observations indicate that the failure cases are generally more compact with their neighbors belonging to ground truth classes and thus hard to attack. We also analyze the failure of ASK-Attack from the visualization perspective. We found that the failure cases usually contain more distinctive features belonging to ground truth classes. For example, the truck looks more stereoscopic.

\paragraph{Societal impact.} The broad motivation of our work is to explore the robustness of kNN-based deep learning models, which has not been thoroughly studied. We believe this goal is highly relevant to the machine learning/artifical intelligence community, and the methods that our paper introduces can be brought to bear on other deep learning/non-parametric learning problems of interest.

\bibliography{reference,refs_adv}
\bibliographystyle{IEEEtran}

\begin{IEEEbiography}[{\includegraphics[width=1in,height=1.25in,clip,keepaspectratio]{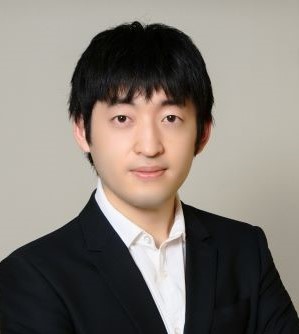}}]{Ren Wang} is currently an assistant professor in the Department of Electrical and Computer Engineering
at the Illinois Institute of Technology. He received bachelor’s degree and master’s degree in Electrical Engineering from Tsinghua University, Beijing, China, in 2013 and 2016. He received his Ph.D. degree in Electrical Engineering from Rensselaer Polytechnic Institute, Troy, NY, USA, in 2020. He was a postdoctoral research fellow in the Department of Electrical Engineering and Computer Science at the University of Michigan. His research interests include Trustworthy Machine Learning, High-Dimensional Data Analysis, Bio-Inspired Machine Learning, and Robustness/Optimization on Smart Grid.
\end{IEEEbiography}

\begin{IEEEbiography}[{\includegraphics[width=1in,height=1.25in,clip,keepaspectratio]{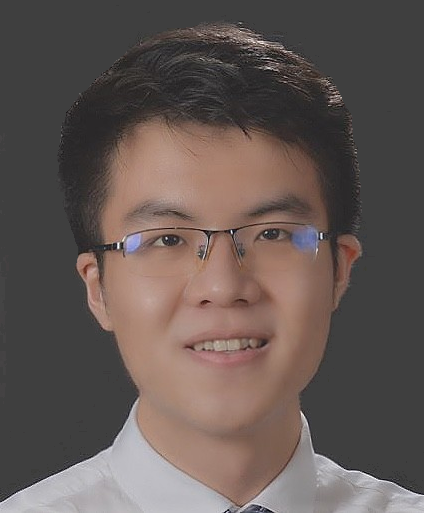}}]{Tianqi Chen} received the B.S. degree in Mathematics from Fudan University, Shanghai, China, in 2019 and the M.S. degree in Statistics from University of Michigan, Ann Arbor, MI, USA, in 2021. He is currently pursuing the Ph.D. degree in Statistics at University of Texas, Austin, TX, USA.
\end{IEEEbiography}

\begin{IEEEbiography}[{\includegraphics[width=1in,height=1.25in,clip,keepaspectratio]{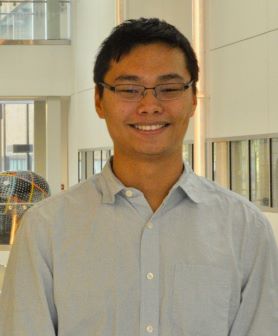}}]{Philip Yao} received his B.S.E degrees in Computer Science and Electrical Engineering from University of Michigan, Ann Arbor, MI, USA, in 2020. He is interested in deep learning research.
\end{IEEEbiography}

\begin{IEEEbiography}[{\includegraphics[width=1in,height=1.25in,clip,keepaspectratio]{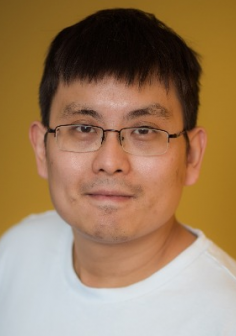}}]{Sijia Liu}
received the Ph.D. degree (with the All University Doctoral Prize) in Electrical and Computer Engineering from Syracuse University, Syracuse, NY, USA, in 2016. He was a Postdoctoral Research Fellow at the University of Michigan, Ann Arbor, in 2016-2017, and a Research Staff Member at the MIT-IBM Watson AI Lab in 2018-2020. His research interests include scalable and trustworthy AI, e.g., adversarial deep learning, optimization theory and methods, computer vision, and computational biology. He received the Best Student Paper Award at the 42nd IEEE International Conference on Acoustics, Speech and Signal Processing (ICASSP). His work has been published at top-tier AI conferences such as NeurIPS, ICML, ICLR, CVPR, and AAAI.
\end{IEEEbiography}

\begin{IEEEbiography}[{\includegraphics[width=1in,height=1.25in,clip,keepaspectratio]{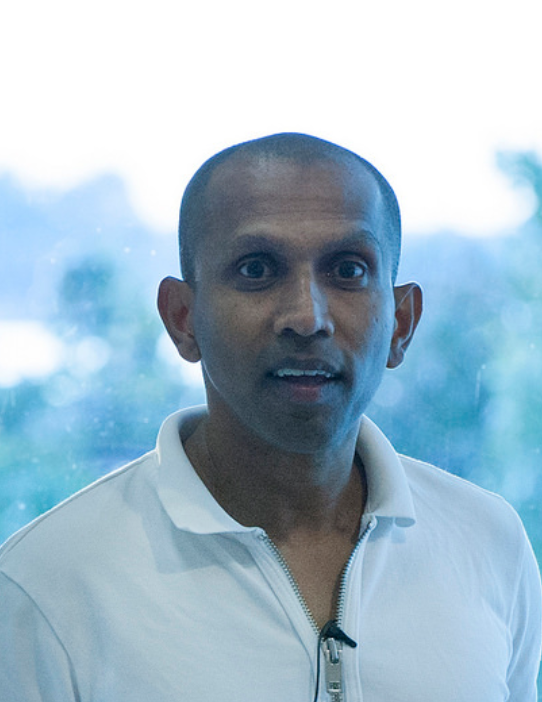}}]{Indika Rajapakse} is currently an Associate Professor of Computational Medicine $\&$ Bioinformatics, in the Medical School, and an Associate Professor of Mathematics at the University of Michigan, Ann Arbor. He is also a member of the Smale Institute. His research is at the interface of biology, engineering and mathematics. His areas include dynamical systems, networks, mathematics of data and cellular reprogramming.
\end{IEEEbiography}

\begin{IEEEbiography}[{\includegraphics[width=1in,height=1.25in,clip,keepaspectratio]{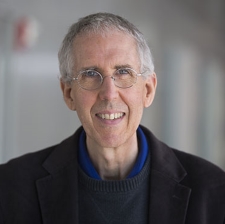}}]{Alfred O. Hero III} (F'97) received the B.S. (summa cum laude) from Boston University (1980) and the Ph.D from Princeton University (1984), both in Electrical Engineering. Since 1984 he has been with the University of Michigan, Ann Arbor, where he is the John H. Holland Distinguished University Professor of Electrical Engineering and Computer Science and the R. Jamison and Betty Williams Professor of Engineering. His primary appointment is in the Department of Electrical Engineering and Computer Science and he also has appointments, by courtesy, in the Department of Biomedical Engineering and the Department of Statistics. His recent research interests are in high dimensional spatio-temporal data, multi-modal data integration, statistical signal processing, and machine learning. Of particular interest are applications to social networks, network security and forensics, computer vision, and personalized health. He is a Section Editor of the SIAM Journal on Mathematics of Data Science and a Senior Editor of the IEEE Journal on Selected Topics in Signal Processing . He is on the editorial board of the Harvard Data Science Review (HDSR) and serves as moderator for the Electrical Engineering and Systems Science category of the arXiv. He was co-General Chair of the 2019 IEEE International Symposium on Information Theory (ISIT) and the 1995 IEEE International Conference on Acoustics, Speech and Signal Processing. He was founding Co-Director of the University’s Michigan Institute for Data Science (MIDAS) (2015-2018). From 2008-2013 he held the Digiteo Chaire d’Excellence at the Ecole Superieure d’Electricite, Gif-sur-Yvette, France. He is a Fellow of the Institute of Electrical and Electronics Engineers (IEEE) and the Society for Industrial and Applied Mathematics (SIAM). Several of his research articles have received best paper awards. He was awarded the University of Michigan Distinguished Faculty Achievement Award (2011), the Stephen S. Attwood Excellence in Engineering Award (2017), and the H. Scott Fogler Award for Professional Leadership and Service (2018). He received the IEEE Signal Processing Society Meritorious Service Award (1998), the IEEE Third Millenium Medal (2000), the IEEE Signal Processing Society Technical Achievement Award (2014), the Society Award from the IEEE Signal Processing Society (2015) and the Fourier Award from the IEEE (2020). He was President of the IEEE Signal Processing Society (2006-2008) and was on the IEEE Board of Directors (2009-2011) where he served as Director of Division IX (Signals and Applications). From 2011 to 2020 he was a member and Chair (2017-2020) of the Committee on Applied and Theoretical Statistics (CATS) of the US National Academies of Science.

\end{IEEEbiography}

\EOD

\end{document}